\documentclass{article} 
\usepackage[final]{colm2026_conference}

\usepackage{microtype}
\usepackage{hyperref}
\usepackage{url}
\usepackage{booktabs}

\usepackage{lineno}

\usepackage{makecell}
\usepackage{multirow} 
\usepackage[most]{tcolorbox}
\usepackage{booktabs} 
\usepackage{amsmath}
\usepackage{listings}
\usepackage[table]{xcolor}
\usepackage[T1]{fontenc}

\definecolor{darkblue}{rgb}{0, 0, 0.5}
\hypersetup{colorlinks=true, citecolor=darkblue, linkcolor=darkblue, urlcolor=darkblue}

\title{Synthesizing Instruction-Tuning Datasets with Contrastive Decoding}

\author{%
\textbf{Tatsuya Ichinose}\textsuperscript{1}\quad
\textbf{Youmi Ma}\textsuperscript{1}\quad
\textbf{Masanari Oi}\textsuperscript{1}\quad  \\
\textbf{Ryuto Koike}\textsuperscript{1}\quad 
\textbf{Naoaki Okazaki}\textsuperscript{1,2,3}\quad
\\
\textsuperscript{1}\; \text{Department of Computer Science, School of Computing, Institute of Science Tokyo}\\
\textsuperscript{2}\; \text{National Institute of Advanced Industrial Science and Technology}\\
\textsuperscript{3}\; \text{Research and Development Center for Large Language Models, NII}\\
\texttt{\{
tatsuya.ichinose@nlp.,
ma.y@,
masanari.oi@nlp.
\}comp.isct.ac.jp} \\
\texttt{\{
ryuto.koike@nlp.,
okazaki@
\}comp.isct.ac.jp} 
}

\newcommand{\methodname}{CoDIT}
\newcommand{\bb}[1]{\textbf{#1}}
\newcommand{\methodfullname}{\bb{Co}ntrastive \bb{D}ecoding for \bb{I}nstruction-\bb{T}uning Dataset}
\newcommand{\methodfullnamelower}{\bb{co}ntrastive \bb{d}ecoding for \bb{i}nstruction-\bb{t}uning dataset}

\begin{document}

\ifcolmsubmission
\linenumbers
\fi

\maketitle

\begin{abstract}
Using responses generated by high-performing large language models (LLMs) for instruction tuning has become a widely adopted approach.
However, the existing literature overlooks a property of LLM-generated responses: they conflate world knowledge acquired during pre-training with instruction-following capabilities acquired during post-training. 
We hypothesize that disentangling the instruction-following capabilities from pre-trained knowledge improves the effectiveness of instruction tuning.
To this end, we propose CoDIT, a method that applies contrastive decoding between a post-trained model and its pre-trained counterpart during response generation.
The method suppresses pre-trained knowledge shared between the two models while amplifying the instruction-following behavior acquired via post-training, resulting in responses that more purely reflect instruction-following capabilities.
Experiment results demonstrate that models trained on datasets constructed via CoDIT consistently outperform those trained on directly generated responses.
Training on our datasets also yields better performance than on existing publicly available instruction-tuning datasets across multiple benchmarks.
Furthermore, we theoretically and empirically show that CoDIT can be interpreted as distilling the chat vector from parameter space to text space, enabling the transfer of instruction-tuning capabilities across models of different architectures.
\footnote{The dataset and code are available at \url{https://huggingface.co/datasets/Tatsuya-Ichinose/CoDIT} and \url{https://github.com/Tatsuya736482/contrastive_decoding_public}, respectively.}

\end{abstract}
\section{Introduction}
Large language models (LLMs) have recently demonstrated strong capabilities in following user instructions. 
These capabilities are acquired through \textit{instruction tuning}, a supervised fine-tuning process in which models learn to generate helpful responses to user instructions~\citep{wei2022finetuned}. 
Instruction tuning requires instruction-response pairs as training data, but manually constructing such data is costly: instructions span diverse domains, necessitating domain experts to ensure response quality.
To address this, existing work has demonstrated that using high-performing LLMs to generate responses can produce effective instruction-tuning datasets at scale~\citep{vicuna2023, xu2024wizardlm, mukherjee2023orcaprogressivelearningcomplex, mitra2023orca2teachingsmall, zhao2024wildchat, zheng2024lmsyschatm, ma2025building}.

However, directly using LLM-synthesized responses introduces a fundamental mismatch in the training objective.
While the goal of instruction tuning is to teach models to follow instructions, LLM-generated responses conflate two distinct components, namely world knowledge acquired during pre-training (hereafter, pre-trained knowledge) and instruction-following abilities acquired during post-training. 
Previous work has shown that such pre-trained knowledge of a teacher cannot be effectively transferred to a student model via supervised fine-tuning~\citep{gudibande2024the}, suggesting that its presence in instruction-tuning data serves as noise rather than a helpful training signal.
Therefore, we hypothesize that suppressing pre-training knowledge during response generation and preserving only instruction-following behavior can yield training data that more purely captures instruction-following ability, leading to more effective instruction tuning.

\begin{figure}
\begin{center}
   \includegraphics[width=0.97\linewidth]{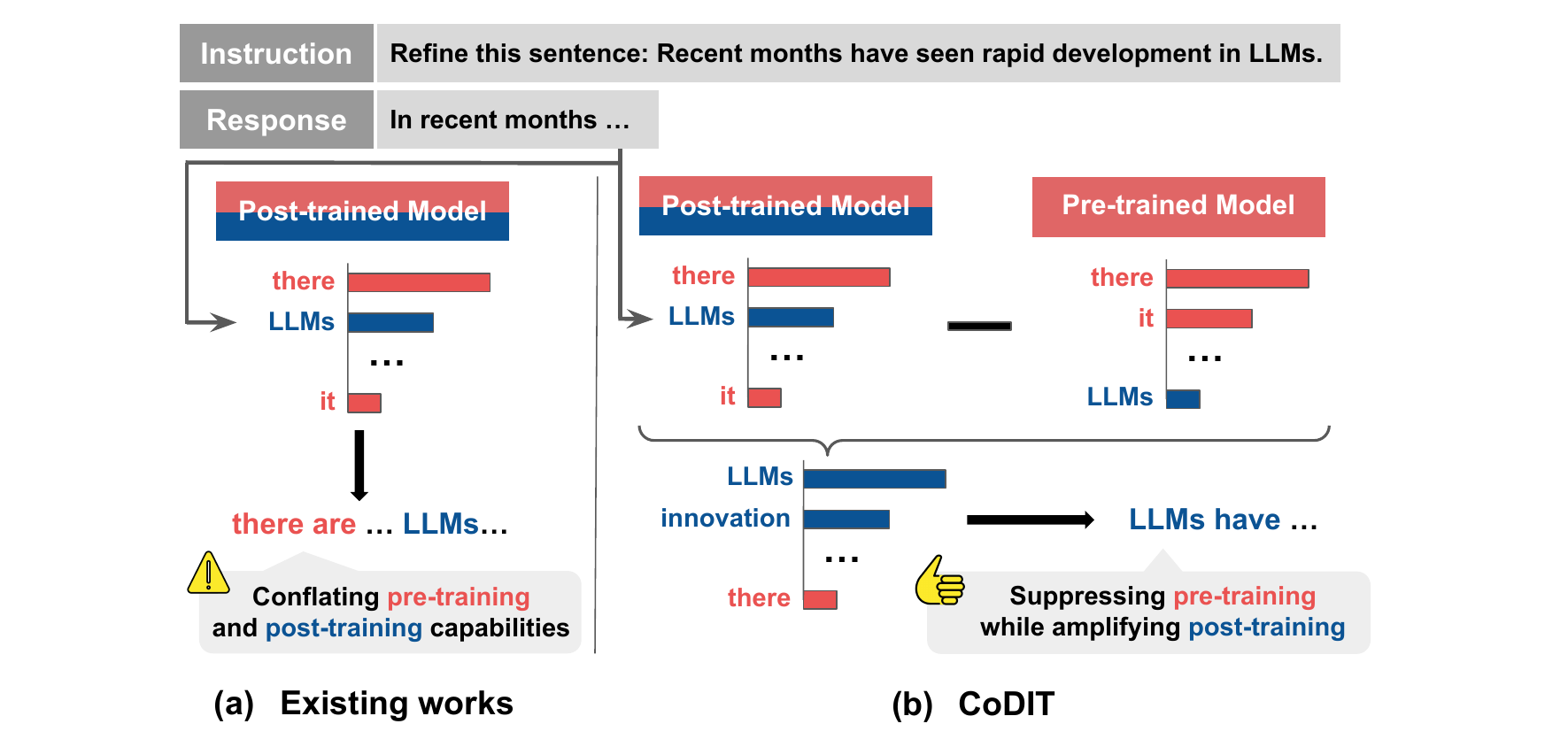}
\end{center}
\caption{Comparison of \methodname~with direct response generation. Existing methods use responses directly generated by the post-trained model (e.g., "In recent months, there are"). In contrast, \methodname~isolates the instruction-following behavior via contrastive decoding, which prioritizes tokens with high likelihood under the post-trained model and low likelihood under the pre-trained model.}
\label{fig:concept}
\end{figure}

This work proposes \bb{\methodname} (\methodfullname), a method that disentangles instruction-following abilities from pre-trained knowledge and suppresses the latter during response generation (Figure \ref{fig:concept}). 
To achieve this, we build on contrastive decoding~\citep{li-etal-2023-contrastive}, a decoding strategy that improves output quality by favoring sequences with high likelihood 
under a large expert model over those under a smaller amateur model.
We instantiate this framework using a post-trained model as the expert and its pre-trained checkpoint as the amateur.
Such a formulation explicitly amplifies the difference between the two models, i.e., the instruction-following capabilities instilled by post-training, while suppressing the pre-trained knowledge they share.

In experiments, we construct datasets using \methodname~with multiple open-weight LLMs -- referred to as \textit{teacher models} -- and evaluate by training separate LLMs -- referred to as \textit{student models} -- on the resulting data.
As measured by WildBench~\citep{lin2025wildbench} and AlpacaEval 2.0~\citep{alpaca_eval}, models trained on \methodname-generated responses consistently outperform those trained on responses directly generated by the teacher model across all nine teacher-student pairs, regardless of model scale or architecture.
This validates our hypothesis that suppressing pre-trained knowledge during response generation yields more effective instruction-tuning data.
Notably, our datasets also outperform existing instruction-tuning datasets~\citep{zhao2024wildchat,xu2025magpie,ma2025building,jiang-etal-2025-instruction}, demonstrating their utility as training resources. 

Furthermore, we theoretically and empirically demonstrate that \methodname~can be interpreted as a text-space distillation of the chat vector \citep{huang-etal-2024-chat}: the difference in model parameters before and after post-training. 
Chat vector operates in the parameter space, limiting its applicability to models with identical architectures.
Our text-space distillation, however, lifts this constraint and enables the transfer of instruction-following capabilities across models of any scale or architecture.

In short, the contributions of this work are as follows: 
(1) We propose~\methodname, a method for constructing instruction-tuning datasets that suppresses pre-trained knowledge during response generation, thereby isolating and emphasizing instruction-following capabilities.
(2) We released datasets constructed via \methodname~that achieve state-of-the-art performance across multiple benchmarks.
(3) We theoretically and empirically demonstrate that \methodname~can be interpreted as a text-space distillation of the chat vector, enabling the transfer of instruction-following capabilities across models of arbitrary architectures.

\section{Methodology}
\label{sec:method_generation}

\subsection{\methodname: \methodfullnamelower}

In this study, we refer to a model that synthesizes responses as the \textbf{teacher model}, and a model trained on the synthesized responses as the \textbf{student model}\footnote{We treat post-trained models as expert models, noting that these models may have undergone additional post-training stages beyond instruction tuning, such as reinforcement learning. An ablation study on the effect of different post-training stages is provided in Section~\ref{sec_chat_vector_distillation} and Appendix~\ref{appendix:impact-of-post-training}.}.
\methodname~applies contrastive decoding~\citep{li-etal-2023-contrastive} to the teacher model during response generation, leveraging the pre-trained model as the amateur and the post-trained model as the expert.

Specifically, given an instruction $x$, the goal is to generate a response $y$.
Let \( \theta_{\text{post}} \) and \( \theta_{\text{pre}} \) denote the parameters of the post-trained and the pre-trained version of the teacher model, respectively.
The response $y$ is decoded token by token, where each token is selected to maximize the difference between the log-probabilities assigned the post-trained and pre-trained model.
For example, the $i$-th token $y_i$ is selected as:

\begin{align}
s(v;x,y_{<i}) & = \log P(v \mid x, y_{<i}; \theta_{\text{post}})
- \log P(v \mid x, y_{<i}; \theta_{\text{pre}}), \label{eq:contrastive_score} \\
y_i & = \arg \max_{v \in \mathcal{V}}s(v;x,y_{<i}), \label{eq:argmax_contrastive_score}
\end{align}
where $s(v;x,y_{<i})$ is the contrastive score, $v$ is a candidate token chosen from the vocabulary $\mathcal{V}$, and $y_{<i}$ is the sequence of previously decoded tokens.
The full response $y$ is obtained by repeating this selection autoregressively until a stop token is reached.

However, as noted in prior work~\citep{li-etal-2023-contrastive}, unrestricted contrastive decoding using \eqref{eq:argmax_contrastive_score} can produce degraded outputs: tokens with very low probability under the pre-trained model may be overemphasized, while tokens with high probabilities under both models may be underemphasized. This can suppress contextually appropriate tokens, degrading the fluency of the generated text. To mitigate this, we follow \citet{li-etal-2023-contrastive} and apply a plausibility constraint, which restricts candidate tokens to those assigned sufficiently high probability by the post-trained model. 
Formally, let $\alpha \in [0,1]$ be a hyperparameter controlling the strictness of the constraint. We define the constrained score as
\begin{equation}
s'(v; x, y_{<i}) =
\begin{cases}
s(v; x, y_{<i}), & 
P(v \mid x, y_{<i}; \theta_{\text{post}})
\ge
\alpha \max_{w \in \mathcal{V}} P(w \mid x, y_{<i}; \theta_{\text{post}}), \\
-\infty, & \text{otherwise.}
\end{cases}
\label{eq:contrastive_decoding}
\end{equation}

\subsection{CoDIT as distilling the chat vector}
\label{sec:method_analysis}
We theoretically show that \methodname~can be interpreted as a text-space distillation of the chat vector~\citep{huang-etal-2024-chat}, a training-free method that transfers post-training abilities between models via parameter arithmetic.

Following~\citet{huang-etal-2024-chat}, the chat vector is defined as the parameter difference between the post-trained and the pre-trained models:
\begin{equation}
\Delta \theta = \theta_{\text{post}} - \theta_{\text{pre}}
\label{eq:chat_vector}
\end{equation}
The chat vector $\Delta \theta$ is considered to capture the capabilities acquired during post-training.

In practice, the parameter shift induced by post-training is typically small relative to the magnitude of the pre-trained parameters ($\|\Delta \theta\| \ll \|\theta_{\text{pre}}\|$).
This motivates a first-order Taylor expansion of $\log P(v \mid x, y_{<i}; \theta)$ around $\theta_{\text{pre}}$, similarly as in~\citet{isonuma2025whats}:
\begin{equation}
\log P(v \mid x, y_{<i}; \theta_{\text{post}})
\approx
\log P(v \mid x, y_{<i}; \theta_{\text{pre}})
+
\Delta\theta^{\top}
\nabla_{\theta} \log P(v \mid x, y_{<i}; \theta_{\text{pre}}).
\end{equation}
Substituting the above equation into Equation~\eqref{eq:contrastive_score}, we approximate the contrastive score as:
\begin{equation}
s(v; x, y_{<i})
\approx
\Delta\theta^{\top}
\nabla_{\theta} \log P(v \mid x, y_{<i}; \theta_{\text{pre}}).
\label{eq:score_approx}
\end{equation}

This shows that the contrastive score is proportional to the inner product between the chat vector $\Delta\theta$ and the gradient of the log-likelihood  $\log P(v \mid x, y_{<i}; \theta_{\text{pre}})$. 
As \methodname~selects $v$ that maximizes the contrastive score $s(v; x, y_{<i})$, the method thus selects the token whose gradient is most aligned with the update direction $\Delta\theta$ induced by post-training. Through this gradient-based alignment, our method effectively distills the teacher model's chat vector into the text space, transferring post-training capabilities across models of arbitrary scale and architecture without parameter-space operations.

\section{Experiments}
\label{sec:main_results}

\subsection{Experimental setup}
\label{sec:experimental_setup}
\paragraph{Teacher Models}
We adopt Qwen3-8B, Qwen3-30B-A3B \citep{yang2025qwen3technicalreport}, and gemma-3-27b-it \citep{gemmateam2025gemma3technicalreport} as our teacher models. 
These models were selected for three reasons: 
(1) both their pre-trained and post-trained checkpoints are publicly available, which is necessary for conducting contrastive decoding; 
(2) they span diverse model families and parameter scales, enabling a comprehensive and robust evaluation; and 
(3) they exhibit strong instruction-following capabilities, ensuring high-quality response generation.

\paragraph{Student Models}
We use Llama-3.1-8B~\citep{grattafiori2024llama3herdmodels}, Qwen3-8B-Base~\citep{yang2025qwen3technicalreport}, and gemma-3-4b-pt~\citep{gemmateam2025gemma3technicalreport} as student models.
The specific training hyperparameters are detailed in Appendix \ref{appendix:generation_hyp}.

\paragraph{Training Dataset}
We synthesize responses by applying \methodname~to the instructions in LMSYS-Chat-1M~\citep{zheng2024lmsyschatm}.
LMSYS-Chat-1M is a dataset consisting of one million human--LLM conversation logs collected from websites such as Chatbot Arena.
Following prior work~\citep{ma2025building}, we preprocess the dataset by removing duplicates, filtering out template-style instructions, and excluding instructions that contain personal information.
After preprocessing, we obtain 250,333 English instructions.
We construct the training dataset by generating a corresponding response for each instruction using \methodname.

\paragraph{Baselines}
We compare against two baselines generated from the same instruction set: (1) \textbf{Vanilla:} A single response generated directly by the teacher model. Most existing approaches 
such as \citet{wang-etal-2023-self-instruct} and \citet{xu2025magpie} 
fall into this category. (2) \textbf{Best-of-N:} For each instruction, we generate five candidate responses and judge the quality of generated responses using gpt-oss-120b~\citep{openai2025gptoss120bgptoss20bmodel}. Specifically, each response is evaluated on a scale from 1 to 10, and the highest-scoring response is selected\footnote{We use a prompt based on WildBench~\citep{lin2025wildbench} for evaluation.}. Detailed experimental settings are provided in Appendix~\ref{appendix:best_of_n}. This setting captures recent methods~\citep{grattafiori2024llama3herdmodels,ma2025building} that incorporate response selection or reranking.

\paragraph{Evaluation Datasets and Metrics}
To evaluate models trained on the synthesized datasets, we employ three LLM-as-a-judge benchmarks: WildBench~\citep{lin2025wildbench}, AlpacaEval 2.0~\citep{dubois2023alpacafarm}, and MT-Bench~\citep{zheng2023judging}.
WildBench provides a comprehensive evaluation using 1,024 samples collected from real-world dialogue logs, each comprising up to five dialogue turns.
We report the WB-Score computed from 10-point ratings produced by GPT-4o (2024-05-13)~\citep{openai2024gpt4technicalreport} and rescaled with 5 as the midpoint (i.e., $S' = (S - 5) \times 2$), with the evaluation prompt described in Appendix~\ref{appendix:wildbench-prompt}.
AlpacaEval 2.0 evaluates responses to 805 instructions via pairwise comparison against GPT-4 Turbo (1106), where the same model also serves as the evaluator.
We report both the Win Rate (WR) and the Length-Controlled Win Rate (LC)~\citep{dubois2024lengthcontrolled}, with the latter one mitigating response-length bias.
MT-Bench consists of 80 high-quality two-turn dialogue samples.
For evaluation, we use GPT-4o (2024-08-06), which scores each response out of 10.
For each instruction, we sample five responses and report the average score.

\subsection{Main results}

\begin{table*}[t]
\small
\centering
\setlength{\tabcolsep}{4pt}
\begin{tabular}{l ccc ccc ccc}
\toprule
\textbf{Teacher} & \multicolumn{3}{c}{\textbf{Qwen3-8B}} & \multicolumn{3}{c}{\textbf{Qwen3-30B-A3B}} & \multicolumn{3}{c}{\textbf{Gemma-3-27B}} \\
\cmidrule(lr){1-1} \cmidrule(lr){2-4} \cmidrule(lr){5-7} \cmidrule(lr){8-10}
Student & Llama & Qwen & Gemma & Llama & Qwen & Gemma & Llama & Qwen & Gemma \\
\midrule

\multicolumn{10}{l}{\textbf{WildBench WB-Score}} \\
\midrule
Vanilla & 43.95 & 60.23 & 30.42 & 42.32 & 57.32 & 30.70 & 48.30 & 60.53 & 34.29 \\
Best-of-N & 44.67 & 60.12 & \textbf{32.82} & 46.44 & 59.30 & 32.66 & 48.97 & 59.90 & 34.96 \\
\rowcolor{gray!10} \methodname & \textbf{48.34} & \textbf{63.18} & 31.84 & \textbf{48.28} & \textbf{60.23} & \textbf{33.07} & \textbf{54.12} & \textbf{62.21} & \textbf{38.89} \\
\midrule

\multicolumn{10}{l}{\textbf{Alpaca Eval 2.0 Win Rates (\%)}} \\
\midrule
Vanilla & 52.55 & 64.76 & 40.42 & 51.78 & 59.47 & 40.36 & 69.51 & 71.67 & 55.12 \\
Best-of-N & 53.35 & 66.54 & 42.93 & 51.79 & 60.81 & 40.60 & 71.46 & 73.27 & 54.88 \\
\rowcolor{gray!15} \methodname & \textbf{58.73} & \textbf{69.45} & \textbf{47.18} & \textbf{56.76} & \textbf{63.79} & \textbf{43.46} & \textbf{75.95} & \textbf{76.26} & \textbf{59.80} \\
\midrule

\multicolumn{10}{l}{\textbf{Alpaca Eval 2.0 Length-Controlled Win Rates (\%)}} \\
\midrule
Vanilla & \textbf{47.92} & 53.46 & 29.25 & 42.03 & 53.10 & 32.47 & 41.43 & 46.23 & 33.04 \\
Best-of-N & 39.73 & 55.82 & 33.09 & 42.70 & 54.85 & 33.78 & 46.13 & 48.84 & 33.03 \\
\rowcolor{gray!15} \methodname & 44.07 & \textbf{56.83} & \textbf{33.43} & \textbf{46.94} & \textbf{55.22} & \textbf{35.50} & \textbf{52.17} & \textbf{51.10} & \textbf{34.57} \\
\midrule

\multicolumn{10}{l}{\textbf{MT-Bench}} \\
\midrule
Vanilla & 73.11 & 84.41 & 63.88 & 72.84 & \textbf{84.39} & 58.67 & \textbf{74.44} & 83.64 & \textbf{67.88} \\
Best-of-N & 71.26 & 84.04 & 61.31 & 72.04 & 83.90 & \textbf{64.42} & 72.91 & \textbf{84.04} & 66.01 \\
\rowcolor{gray!15} \methodname & \textbf{75.39} & \textbf{85.25} & \textbf{64.96} & \textbf{73.72} & 84.17 & 62.82 & 74.33 & 83.12 & 67.14 \\

\bottomrule
\end{tabular}
\caption{Evaluation scores of instruction-following performance for models trained on instruction-tuning datasets constructed by the proposed method and baselines. We confirm that \methodname~yields consistently better instruction-tuning effectiveness than the baselines.}
\label{tab:main_results}
\end{table*}

Table~\ref{tab:main_results} presents the results for each experimental setting.
On WildBench and AlpacaEval 2.0, \methodname~yields consistent gains over both Vanilla and Best-of-N baselines. On WildBench, \methodname~achieves the best WB-Score in 8 out of 9 configurations and improves the overall average by a substantial margin. 
The advantage of \methodname~is even more pronounced on AlpacaEval 2.0, where \methodname~achieves the highest win rate across all teacher--student pairs. 
In contrast, the gains on MT-Bench are more modest. We attribute this to the small sample size of the benchmark, a limitation also pointed out in prior work \citep{lin2025wildbench}.
Overall, across multiple settings, models trained on \methodname-generated responses outperform those trained on directly-generated ones, validating our hypothesis that \textbf{suppressing pre-trained knowledge during response generation yields more effective instruction-tuning data}.

\begin{table}[t]
\small
\centering
\begin{tabular}{lcll}
\toprule
\textbf{Dataset} & \textbf{\# Instances} & \textbf{Instruction} & \textbf{Response} \\
\midrule
WildChat & 778,380 & WildChat & GPT-3.5, GPT-4 \\
Llama-3.1-LMSYS & 453,737 & LMSYS-Chat-1M & Llama-3.1-405B-Instruct \\
Gemma-2-LMSYS & 453,861 & LMSYS-Chat-1M & gemma-2-27b-it \\
Magpie-Pro-300K-Filtered & 299,912 & Llama-3-70B-Instruct & Llama-3-70B-Instruct \\
WebR-Basic  & 87,490 & \multicolumn{2}{l}{Rewrite web text using Llama-3-70B-Instruct}  \\
WebR-Pro & 87,498 & \multicolumn{2}{l}{Rewrite web text using GPT-4o-mini} \\
\midrule 
\methodname-Gemma3 & 250,333 & LMSYS-Chat-1M & gemma-3-27b-it \\
\methodname-Qwen3-8B & 250,333 & LMSYS-Chat-1M & Qwen3-8B \\
\methodname-Qwen3-30B & 250,333 & LMSYS-Chat-1M & Qwen3-30B-A3B \\
\bottomrule
\end{tabular}
\caption{Summary of instruction-tuning datasets. ``-Chat-1M-Synth'' is omitted from Llama-3.1-LMSYS and Gemma-2-LMSYS for brevity.}
\label{tab:dataset_summary}
\end{table}

\begin{table}[t]
\small
\centering
\tabcolsep 1pt
\begin{tabular}{l cccc  cccc}
\toprule
& 
\multicolumn{8}{c}{\textbf{Student Model}}  \\
& 
\multicolumn{4}{c}{\textbf{Llama-3.1-8B}} &
\multicolumn{4}{c}{\textbf{Qwen3-8B-Base}} \\
\cmidrule(lr){2-5}\cmidrule(lr){6-9}
\multirow{2}{*}{\textbf{Dataset}} &
\makecell[c]{Wild\\Bench} &
\multicolumn{2}{c}{\makecell[c]{Alpaca\\Eval2.0}} &
\makecell[c]{MT\\Bench} &
\makecell[c]{Wild\\Bench}  &
\multicolumn{2}{c}{\makecell[c]{Alpaca\\Eval2.0}} &
\makecell[c]{MT\\Bench}  \\
\cmidrule(lr){3-4}\cmidrule(lr){7-8}
& 
 \makecell[c]{WB\\Score} &
 LC &
 WR &
 AVG & 
 \makecell[c]{WB\\Score} &
 LC &
 WR &
 AVG \\
\midrule
WildChat  
& 21.69 & 15.07 & 9.54 & 62.24 & 29.99 & 23.66 & 14.49 & 70.04 \\
Llama-3.1-LMSYS-Chat-1M-Synth 
& 30.94 & 27.81 &	27.84 & 72.11 & 42.05 & 34.58 & 31.32 & 80.05\\
Gemma-2-LMSYS-Chat-1M-Synth
& 36.20 & 40.74 &31.87 & 69.51 & 46.84 & 45.52 &33.80 & 74.56\\
Magpie-Pro-300K-Filtered & 26.85 & 24.77 & 31.64 & 63.15 & 42.91 & 34.81 & 39.49 & 74.05 \\
WebR-Basic & 26.57 & 20.30 &22.62 & 64.31 & 41.20 & 37.17 &38.58 & 75.43 \\
WebR-Pro & 37.19 & 32.65 & 34.46 & 70.00 & 54.40 & 46.22 & 45.06 & 82.72 \\
\hline
\rowcolor{gray!15}
\methodname-Gemma3
& \bb{54.12}  & \bb{52.73} & \bb{72.85} & 74.33 &  62.21 & 47.98 & \bb{71.41} & 83.13  \\
\rowcolor{gray!15}
\methodname-Qwen3-8B
& 48.34 & 42.75 & 55.30 & \bb{75.39} &  \bb{63.18} & \bb{54.25} & 65.62 & \bb{85.25} \\
\rowcolor{gray!15}
\methodname-Qwen3-30B
& 48.28  & 46.03 & 54.71 & 73.73 & 60.23  & 52.74 &  60.36 & 84.18  \\
\bottomrule
\end{tabular}
\caption{Performance comparison between publicly available instruction-tuning datasets and our synthesized datasets. Our dataset outperforms existing datasets.}
\label{tab:results2}
\end{table}

\subsection{Comparison with existing instruction-tuning datasets}

To evaluate the quality of datasets constructed using \methodname~related to existing resources, we compare models trained on our datasets against those trained on publicly available instruction-tuning datasets, including WildChat \citep{zhao2024wildchat}, Llama-3.1-LMSYS-Chat-1M-Synth, Gemma-2-LMSYS-Chat-1M-Synth \citep{ma2025building}, Magpie-Pro-300K-Filtered~\citep{xu2025magpie}, WebR-Basic, and WebR-Pro~\citep{jiang-etal-2025-instruction}.
Table~\ref{tab:dataset_summary} summarizes these datasets and experiment results are reported in Table~\ref{tab:results2}~\footnote{For the AlpacaEval 2.0 evaluation, we utilized GPT-4 Turbo (2024-04-09) as the automated judge.}.
Across all benchmarks and for all student models, training on datasets constructed by \methodname~consistently outperforms training on existing datasets. These results demonstrate that \textbf{datasets constructed using \methodname~serves as effective language resources for instruction tuning}.

\section{Analysis}
\subsection{Do performance gains stem from improved response quality?}
\label{sec:score_distribution}

\begin{figure}[t]
\begin{center}
   \includegraphics[width=0.67\linewidth]{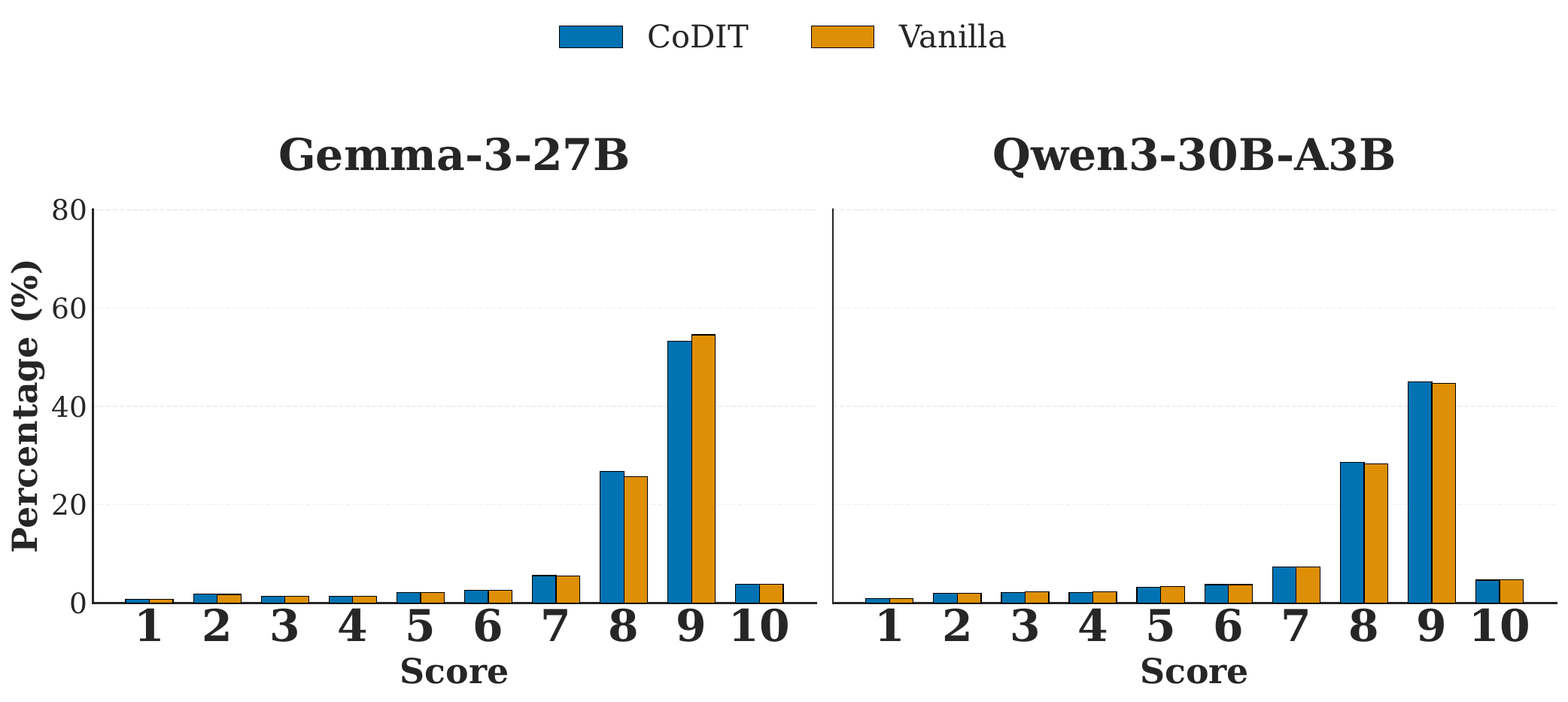}
\end{center}
\caption{Score distribution of the synthetic dataset evaluated by gpt-oss-120b on a scale of 1 to 10. A comparison between responses generated with and without \methodname~reveals that \methodname~does not affect output quality.
}
\label{fig:score-distribution}
\end{figure}
Since contrastive decoding is commonly used to improve text quality~\citep{liu-etal-2021-dexperts,li-etal-2023-contrastive, chuang2024dola}, one may question whether the performance gains of \methodname~stem from improved response quality. In this section, we rule out this alternative explanation and confirm that the observed improvements are attributed to the distillation of instruction-following capabilities.

Specifically, we utilize gpt-oss-120b as an LLM-as-a-judge to assess the quality of responses generated directly and using \methodname, employing an evaluation prompt adapted from WildBench as in Appendix \ref{appendix:best_of_n}. 
As illustrated in Figure~\ref{fig:score-distribution}, the evaluation reveals that the text quality of the responses generated by \methodname~stays at a similar level as those generated directly by the teacher model (Vanilla). 
This similarity in score distributions strongly indicates that \textbf{the effectiveness of \methodname~does not stem from simply generating higher-quality text, but rather from successfully isolating and transferring the targeted instruction-following behaviors}. 
These findings redefine contrastive decoding as a novel framework for isolating specific latent capabilities within LLMs.

\subsection{Does \methodname~better distill the chat vector?}
\label{sec_chat_vector_distillation}
\begin{figure}[t]
\begin{center}
   \includegraphics[width=0.85\linewidth]{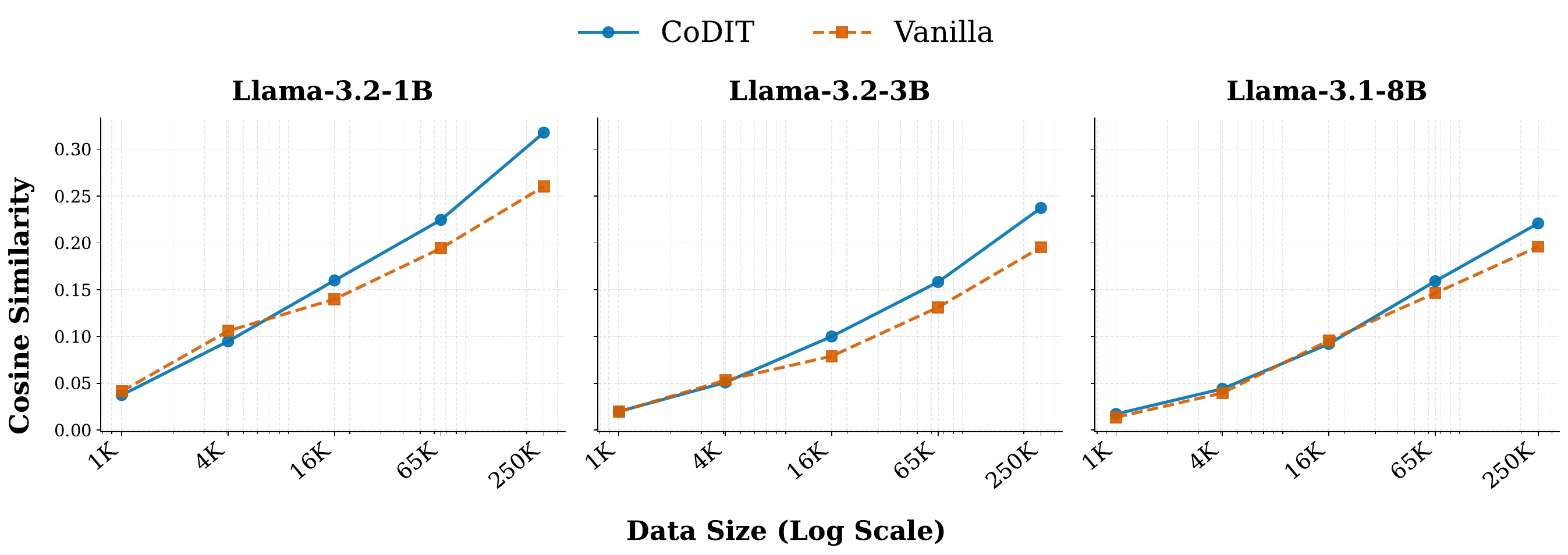}
\end{center}
\caption{Cosine similarity between the teacher's chat vector $\Delta \theta$ and the update vector $\Delta \theta'$. 
Compared to Vanilla, \methodname~exhibits consistently higher similarity as the data scale grows, with the performance gap widening at larger scales. This indicates that \methodname~more effectively distills the teacher model's chat vector into the text-space.}
\label{fig:chat_vector_sim}
\end{figure}

As discussed in Section \ref{sec:method_analysis}, \methodname~conceptually acts as distilling the teacher model's chat vector~\citep{huang-etal-2024-chat} into the text space. To empirically verify this, we examine whether fine-tuning on datasets constructed using \methodname~induces parameter updates that align with the teacher model's chat vector, measured by cosine similarity between the two vectors.

\paragraph{Teacher Model's Chat Vector}
We extract the teacher model's chat vector
$
\Delta \theta = \theta_{\mathrm{post}} - \theta_{\mathrm{pre}}
$ (Equation~\eqref{eq:chat_vector})
which represents the parameter shift induced by post-training. Here, $\theta_{\mathrm{pre}}$ and $\theta_{\mathrm{post}}$ denote the parameters of the publicly released pre-trained and post-trained checkpoints of the teacher model, respectively.

\paragraph{Parameter Updates Induced by \methodname}
We instruction-tune the teacher's pre-trained checkpoint $\theta_{\mathrm{pre}}$ on datasets constructed by applying \methodname~to $\theta_{\text{post
}}$ and $\theta_{\text{pre}}$, yielding model $\theta_{\mathrm{post}}'$.
We then extract the induced parameter update $\Delta \theta' = \theta_{\mathrm{post}}' - \theta_{\mathrm{pre}}$ and compute its cosine similarity with the teacher's chat vector $\Delta \theta$, from which way we quantify how accurately the chat vector is distilled.

\paragraph{Models}
To examine the effect of model scale, we include Llama-3.2-1B-Instruct, Llama-3.2-3B-Instruct, and Llama-3.1-8B-Instruct and their corresponding pre-trained checkpoints~\citep{grattafiori2024llama3herdmodels}
\footnote{We provide more results on additional model families in Appendix \ref{appendix:additional_chat_vector_distillation}.}.

\paragraph{Training Data} We adopt the same instruction set as described in Section~\ref{sec:experimental_setup} and prepare two response variants: one generated using \methodname~and one using raw outputs from the post-trained models as a baseline (noted as Vanilla). 
In addition to the full dataset of 250,333 samples, we experiment on randomly down-sampled subsets of 1,024, 4,096, 16,384, and 65,536 samples (noted as 1k, 4k, 16k, and 64k) to investigate the effect of data volume.

\paragraph{Results} Figure~\ref{fig:chat_vector_sim} shows the cosine similarity between the teacher's chat vector and the parameter updates induced by fine-tuning on \methodname-generated responses.
\methodname~consistently achieves higher similarity with the teacher model’s chat vector than the baseline, confirming that it more effectively distills the chat vector into the text space.
This gap exists for all evaluated models and widens with increasing data volume.
These results demonstrate that \textbf{\methodname~yields a closer approximation of the teacher model's chat vector}, providing empirical support for the theoretical connection established in Section~\ref{sec:method_analysis}.

\subsection{Does reinforcement learning hinder the effectiveness of \methodname?}

While \methodname~uses the post-trained checkpoint as the expert and the pre-trained checkpoint as the amateur, publicly released post-trained models typically undergo multiple stages as a combination of instruction tuning (IT) and reinforcement learning~(RL,~\citet{grattafiori2024llama3herdmodels, yang2025qwen3technicalreport}). 
Here, we investigate whether the composition of these stages hinders the effectiveness of \methodname.

\paragraph{Models} 
We experiment with Olmo-3~\citep{olmo2025olmo3}, a representative model family that provides checkpoints at different post-training stages. Specifically, we denote the pre-trained checkpoint as $\theta_{\mathrm{pre}}$ (Olmo-3-1025-7B). We then contrast this with the instruction-tuned checkpoint $\theta_{\mathrm{inst}}$ (Olmo-3-7B-Instruct-SFT) and the fully post-trained checkpoint $\theta_{\mathrm{post}}$ (Olmo-3-7B-Instruct), which includes reinforcement learning.

\paragraph{Evaluation} Following Section~\ref{sec_chat_vector_distillation}, we compute the cosine similarity between each teacher checkpoint's chat vector ($\Delta \theta_{\mathrm{inst}}$, $\Delta \theta_{\mathrm{post}}$) and the corresponding parameter update induced by fine-tuning on its synthesized responses (
$\Delta \theta'_{\mathrm{inst}}$, $\Delta \theta'_{\mathrm{post}}$).

\begin{table}[t]
\small
\centering
\begin{tabular}{lccc}
\toprule
& & \multicolumn{2}{c}{\textbf{Similarity with Teacher's chat vector}} \\ \cmidrule(lr){3-4}
\textbf{Teacher Model} & \textbf{Response Generation}  & \textbf{IT} ($\Delta \theta_{\mathrm{inst}}$) & \textbf{IT+RL} ($\Delta \theta_{\mathrm{post}}$) \\
\midrule
\multirow{2}{*}{IT ($\Delta \theta'_{\mathrm{inst}}$)} & Vanilla & 0.1121 & 0.1120 \\
& \methodname & 0.1331 & 0.1331 \\
\midrule
\multirow{2}{*}{IT+RL ($\Delta \theta'_{\mathrm{post}}$)} & Vanilla & 0.1323 & 0.1323 \\
& \methodname & \textbf{0.1493} & \textbf{0.1494} \\
\bottomrule
\end{tabular}
\caption{Cosine similarity between parameter update induced by teacher's synthesized responses at different stages and the teacher's chat vectors. Our results suggest that RL models are more effective at distilling chat vectors.}
\label{tab:chat_vector_similarity_stages}
\end{table}

\paragraph{Results} As shown in Table~\ref{tab:chat_vector_similarity_stages}, training with \methodname-generated responses consistently yields higher cosine similarity with the teacher model's chat vectors than the baseline, regardless of whether the instruction-tuned or fully post-trained checkpoint is used for data synthesis.
Notably, responses generated by the fully post-trained model more effectively distill the instruction-tuned chat vector ($\Delta \theta_{\mathrm{inst}}$) than those generated by the instruction-tuned checkpoint itself (0.1493 v.s. 0.1331), suggesting that \textbf{RL does not hinder but rather amplifies the effectiveness of \methodname}.
We attribute this to RL sharpening the model's output distribution and eliciting latent instruction-following capabilities~\citep{liu2025understanding,zhao2025echo,yue2025does,matsutani2026rl}, thereby strengthening the directional alignment with the instruction-tuning update. 
These results indicate that the RL stage in standard post-training pipelines benefits \methodname, demonstrating the broad compatibility of our proposed method.

\subsection{Does instruction overlap inflate \methodname's performance?}

\begin{table}[t]
\centering
\small
\begin{tabular}{l r ccc}
\toprule
\textbf{Evaluation Set} & \textbf{Size} & \textbf{Exact (\%)} & \textbf{$n$-gram (\%)} & \textbf{Cosine Similarity (\%)} \\
\midrule
MT-Bench & 80 & 32.50 & 51.25 & 46.25 \\
AlpacaEval 2.0 & 805 & 23.73 & 39.01 & 32.05 \\
WildBench & 1{,}024 & 0.20 & 2.05 & 0.68 \\
\bottomrule
\end{tabular}
\caption{Overlap between LMSYS-Chat-1M training instructions and evaluation prompts.}
\label{tab:contamination_overlap}
\end{table}

\begin{table}[t]
\centering
\small
\begin{tabular}{l cc c}
\toprule
\multirow{2}{*}{\textbf{Dataset}} & \multicolumn{2}{c}{\textbf{AlpacaEval 2.0}} & \textbf{WildBench} \\
\cmidrule(lr){2-3} \cmidrule(lr){4-4}
 & LC & WR & WB-Score \\
\midrule
WildChat & 13.42 & 8.46 & 19.81 \\
Gemma-2-LMSYS-Chat-1M-Synth & 37.52 & 30.13 & 33.75 \\
Magpie-Pro-300K-Filtered & 25.21 & 32.20 & 26.52 \\
WebR-Pro & 32.71 & 35.93 & 39.39 \\
\midrule
\rowcolor{gray!10}
\methodname-Gemma3 & \bb{47.34} & \bb{69.60} & \bb{52.05} \\
\rowcolor{gray!10}
\methodname-Qwen3-8B & 41.73 & 53.49 & 46.82 \\
\bottomrule
\end{tabular}
\caption{Performance comparison after removing overlap between LMSYS-Chat-1M and evaluation benchmarks.}
\label{tab:contamination_results}
\end{table}

LMSYS-Chat-1M~\citep{zheng2024lmsyschatm} instructions, which we use to construct our datasets, may overlap with evaluation prompts in MT-Bench, AlpacaEval 2.0, and WildBench. This raises a natural question: Does this instruction overlap give \methodname~an unfair advantage?

\paragraph{Measuring Overlap}
To answer this question, we first quantify the extent of instruction overlap.
We measure exact-match, 13-gram character-level, and semantic cosine similarity using all-mpnet-base-v2~\footnote{\url{https://huggingface.co/sentence-transformers/all-mpnet-base-v2}} between LMSYS-Chat-1M instructions and each evaluation set.
For multi-turn benchmarks, we use the maximum turn-level similarity.

Table~\ref{tab:contamination_overlap} reports the percentage of LMSYS-Chat-1M instructions with similarity $\ge 0.8$ to at least one evaluation prompt, measured by $n$-gram overlap and the cosine similarity of sentence embeddings.
As shown in Table~\ref{tab:contamination_overlap}, 32.50\% of the instructions in MT-Bench and 23.73\% of the instructions in AlpacaEval 2.0 overlap with those in LMSYS-Chat-1M, which is non-negligible.

\paragraph{Controlling for Overlap}
To directly test whether this overlap inflates \methodname's performance, we remove instructions similar to those in MT-Bench, AlpacaEval 2.0, or WildBench and re-train the model on the filtered dataset. Specifically, we treat any instruction with a character-level 13-gram similarity of at least 0.8 to an evaluation instruction as similar and filter it out. We then train Llama-3.1-8B on 80K instances from each filtered dataset.

\paragraph{Results}
As shown in Table~\ref{tab:contamination_results}, \methodname-generated datasets still outperform all baselines even after filtering. This confirms that \textbf{\methodname's performance gains are not attributable to instruction overlap with the evaluation sets}, but to the superiority of the method itself.

\section{Related work}
\subsection{Instruction tuning datasets}
The quality and diversity of instruction-tuning datasets are pivotal to the post-training performance of Large Language Models (LLMs). Existing approaches to constructing these datasets can be broadly categorized into two groups: data curation and data synthesis.

\paragraph{Data Curation}
Early approaches focused on reshaping existing NLP benchmarks into instruction-response formats using manual templates~\citep{wang2022supernaturalinstructionsgeneralizationdeclarativeinstructions}.
More recently, the focus has shifted toward leveraging vast amounts of web-scale text \citep{yue2024mammoth, li2024selfalignment, nguyen-etal-2024-better, jiang-etal-2025-instruction}, while others emphasize data filtering to extract a high-quality subset from massive instruction datasets 
\citep{ivison2023camelschangingclimateenhancing, zhang2023modsmodelorienteddata, liu2024what,  li-etal-2024-quantity,zhang2025the}.
Despite their effectiveness, these methods remain inherently limited by the quality and distribution of existing data.

\paragraph{Data Synthesis} 
Given the high cost of manually constructing instruction-tuning datasets, recent work has shifted toward using LLMs to synthesize data.
One common strategy is to collect real-world user queries from chat logs \citep{vicuna2023,zheng2024lmsyschatm,zhao2024wildchat}.
Another line of work generates diverse instructions from a small seed collection \citep{wang-etal-2023-self-instruct,alpaca,xu2024wizardlm,wang-etal-2024-codeclm}. More recently, Magpie \citep{xu2025magpie} demonstrated that complex instructions can be synthesized by prompting LLMs with pre-query templates.
While these methods effectively enhance the diversity and complexity of instructions \citep{zhao-etal-2024-tree}, they directly use LLM outputs as responses as they are.
Furthermore, recent efforts have explored synthesizing datasets tailored to specific student models~\citep{hu-etal-2025-nile, li-etal-2025-mosaic}, though such approaches often lack generalizability across model families. 
In contrast, our method optimizes the response generation process itself and is agnostic to both the instruction source and the target student model, making it complementary to existing instruction-generation frameworks and broadly applicable across model families.

\subsection{Contrastive decoding}

Contrastive decoding \citep{li-etal-2023-contrastive,chang-etal-2024-explaining} is a decoding strategy that leverages an amateur model to improve the output quality of an expert model~\citep{obrien2024contrastive}. Prior work has applied this framework to a range of goals, including reducing toxicity \citep{liu-etal-2021-dexperts,lu2025unidetox}, mitigating hallucinations \citep{chuang2024dola,shi-etal-2024-trusting,sanchez2024stay,zhang-etal-2025-alleviating}, aligning outputs with human preferences \citep{gao2024linear}, and using weaker models to guide and improve stronger ones~\citep{liu2024tuning,jiang2026contrastive}. Beyond output quality, contrastive decoding has also been applied to evaluation \citep{lu2024opendomain} and dataset distillation \citep{isonuma2025whats}.
Unlike prior work, which treats contrastive decoding purely as an inference-time technique, our work leverages it as a principled mechanism for disentangling instruction-following capabilities from pre-trained knowledge. By treating the pre-trained/post-trained versions of a model as the amateur/expert pair, we show that the resulting text-space outputs distill the chat vector~\citep{huang-etal-2024-chat}. Our approach enables capability transfer across models of arbitrary scale and architecture without parameter-space operations.

\section{Conclusion}
We presented \methodname, a novel approach that uses contrastive decoding between a pre-trained and a post-trained model to synthesize training data for instruction tuning.
Experiments across nine teacher-student pairs demonstrated that \methodname-constructed datasets consistently outperform both directly generated responses and existing open-source instruction-tuning datasets, achieving state-of-the-art performance across diverse model families and scales.
We theoretically and empirically showed that \methodname~can be interpreted as a distillation of the chat vector from the parameter space to the text space, enabling the transfer of instruction-tuning abilities between models of arbitrary scale and architecture.

For future work, we plan to extend \methodname~to more complex and specialized settings, such as reasoning tasks, multi-turn conversations, and safety alignment. We believe that selectively amplifying specific capabilities via contrastive decoding provides a principled, general framework for generating high-quality training data across a wide range of LLM development scenarios.

\newpage

\section*{Acknowledgments}
This work was supported by JSPS KAKENHI Grant Number 25H01137.
These research results were obtained from the commissioned research (No.22501) by National Institute of Information and Communications Technology (NICT) , Japan.
This study was carried out using the TSUBAME4.0 supercomputer at Institute of Science Tokyo.
We thank Sakae Mizuki for his valuable suggestions on this research.

\section*{Ethics Statement}
This work focuses on evaluating the usefulness of generated responses in our experimental setting. Therefore, the harmfulness of the responses is not considered.
All instructions in our dataset are derived from LMSYS-Chat-1M~\citep{zheng2024lmsyschatm}, where tokens that could reveal Personally Identifiable Information (PII) are masked out.


\bibliography{colm2026_conference}

@inproceedings{li-etal-2023-contrastive,
    title = "Contrastive Decoding: Open-ended Text Generation as Optimization",
    author = "Li, Xiang Lisa  and
      Holtzman, Ari  and
      Fried, Daniel  and
      Liang, Percy  and
      Eisner, Jason  and
      Hashimoto, Tatsunori  and
      Zettlemoyer, Luke  and
      Lewis, Mike",
    booktitle = "Proceedings of the 61st Annual Meeting of the Association for Computational Linguistics (ACL)",
    year = "2023",
    publisher = "Association for Computational Linguistics",
    pages = "12286--12312",
}

@inproceedings{
    wei2022finetuned,
    title={Finetuned Language Models are Zero-Shot Learners},
    author={Jason Wei and Maarten Bosma and Vincent Zhao and Kelvin Guu and Adams Wei Yu and Brian Lester and Nan Du and Andrew M. Dai and Quoc V Le},
    booktitle={The Tenth International Conference on Learning Representations (ICLR)},
    year={2022},
    url={https://openreview.net/forum?id=gEZrGCozdqR}
}

@inproceedings{
    zhang2025the,
    title={The Best Instruction-Tuning Data are Those That Fit},
    author={Dylan Zhang and Qirun Dai and Hao Peng},
    booktitle={The Thirty-ninth Annual Conference on Neural Information     Processing Systems (NeurIPS)},
    year={2025},
    url={https://openreview.net/forum?id=4jFSekBaDT}
}

@inproceedings{
    xu2024wizardlm,
    title={Wizard{LM}: Empowering Large Pre-Trained Language Models to Follow Complex   Instructions},
    author={Can Xu and Qingfeng Sun and Kai Zheng and Xiubo Geng and Pu Zhao and    Jiazhan Feng and Chongyang Tao and Qingwei Lin and Daxin Jiang},
    booktitle={The Twelfth International Conference on Learning Representations (ICLR)},
    year={2024},
    url={https://openreview.net/forum?id=CfXh93NDgH}
}

@misc{mukherjee2023orcaprogressivelearningcomplex,
      title={Orca: Progressive Learning from Complex Explanation Traces of GPT-4}, 
      author={Subhabrata Mukherjee and Arindam Mitra and Ganesh Jawahar and Sahaj Agarwal and Hamid Palangi and Ahmed Awadallah},
      year={2023},
      eprint={2306.02707},
      archivePrefix={arXiv},
      primaryClass={cs.CL},
      url={https://arxiv.org/abs/2306.02707}, 
}

@misc{mitra2023orca2teachingsmall,
      title={Orca 2: Teaching Small Language Models How to Reason}, 
      author={Arindam Mitra and Luciano Del Corro and Shweti Mahajan and Andres Codas and Clarisse Simoes and Sahaj Agarwal and Xuxi Chen and Anastasia Razdaibiedina and Erik Jones and Kriti Aggarwal and Hamid Palangi and Guoqing Zheng and Corby Rosset and Hamed Khanpour and Ahmed Awadallah},
      year={2023},
      eprint={2311.11045},
      archivePrefix={arXiv},
      primaryClass={cs.AI},
      url={https://arxiv.org/abs/2311.11045}, 
}

@misc{yang2025qwen3technicalreport,
      title={Qwen3 Technical Report}, 
      author={An Yang 
      and Anfeng Li 
      and Baosong Yang 
      and Beichen Zhang 
      and Binyuan Hui 
      and Bo Zheng 
      and Bowen Yu 
      and Chang Gao 
      and Chengen Huang 
      and Chenxu Lv 
      and Chujie Zheng 
      and Dayiheng Liu 
      and Fan Zhou 
      and Fei Huang 
      and Feng Hu 
      and others},
      year={2025},
      eprint={2505.09388},
      archivePrefix={arXiv},
      primaryClass={cs.CL},
      url={https://arxiv.org/abs/2505.09388}, 
}

@misc{vicuna2023,
    title = {Vicuna: An Open-Source Chatbot Impressing {GPT-4} with 90\%* {ChatGPT} Quality},
    url = {https://lmsys.org/blog/2023-03-30-vicuna/},
    note =      "\url{https://lmsys.org/blog/2023-03-30-vicuna/}",
    author = {Chiang, Wei-Lin and Li, Zhuohan and Lin, Zi and Sheng, Ying and Wu, Zhanghao and Zhang, Hao and Zheng, Lianmin and Zhuang, Siyuan and Zhuang, Yonghao and Gonzalez, Joseph E. and Stoica, Ion and Xing, Eric P.},
    year = {2023}
}

@inproceedings{
zhao2024wildchat,
title={{WildChat}: 1{M} {ChatGPT} Interaction Logs in the Wild},
author={Wenting Zhao and Xiang Ren and Jack Hessel and Claire Cardie and Yejin Choi and Yuntian Deng},
booktitle={The Twelfth International Conference on Learning Representations (ICLR)},
year={2024},
url={https://openreview.net/forum?id=Bl8u7ZRlbM}
}

@inproceedings{
ma2025building,
title={Building Instruction-Tuning Datasets from Human-Written Instructions with Open-Weight Large Language Models},
author={Youmi Ma and Sakae Mizuki and Kazuki Fujii and Taishi Nakamura and Masanari Ohi and Hinari Shimada and Taihei Shiotani and Koshiro Saito and Koki Maeda and Kakeru Hattori and Takumi Okamoto and Shigeki Ishida and Rio Yokota and Hiroya Takamura and Naoaki Okazaki},
booktitle={Second Conference on Language Modeling (COLM)},
year={2025},
url={https://openreview.net/forum?id=6vTv9M9ZAA}
}

@inproceedings{
zheng2023judging,
title={Judging {LLM}-as-a-Judge with {MT}-{B}ench and {Chatbot Arena}},
author={Lianmin Zheng and Wei-Lin Chiang and Ying Sheng and Siyuan Zhuang and Zhanghao Wu and Yonghao Zhuang and Zi Lin and Zhuohan Li and Dacheng Li and Eric Xing and Hao Zhang and Joseph E. Gonzalez and Ion Stoica},
booktitle={Thirty-seventh Conference on Neural Information Processing Systems Datasets and Benchmarks Track},
year={2023},
url={https://openreview.net/forum?id=uccHPGDlao}
}

@inproceedings{
dubois2023alpacafarm,
title={{AlpacaFarm}: A Simulation Framework for Methods that Learn from Human Feedback},
author={Yann Dubois and Xuechen Li and Rohan Taori and Tianyi Zhang and Ishaan Gulrajani and Jimmy Ba and Carlos Guestrin and Percy Liang and Tatsunori Hashimoto},
booktitle={Thirty-seventh Conference on Neural Information Processing Systems (NeurIPS)},
year={2023},
url={https://openreview.net/forum?id=4hturzLcKX}
}

@inproceedings{
lin2025wildbench,
title={{WildBench}: Benchmarking {LLM}s with Challenging Tasks from Real Users in the Wild},
author={Bill Yuchen Lin and Yuntian Deng and Khyathi Chandu and Abhilasha Ravichander and Valentina Pyatkin and Nouha Dziri and Ronan Le Bras and Yejin Choi},
booktitle={The Thirteenth International Conference on Learning Representations (ICLR)},
year={2025},
url={https://openreview.net/forum?id=MKEHCx25xp}
}

@inproceedings{
dubois2024lengthcontrolled,
title={{Length-Controlled AlpacaEval}: A Simple Debiasing of Automatic Evaluators},
author={Yann Dubois and Percy Liang and Tatsunori Hashimoto},
booktitle={First Conference on Language Modeling (COLM)},
year={2024},
url={https://openreview.net/forum?id=CybBmzWBX0}
}

@inproceedings{rajbhandari2020zeromemoryoptimizationstraining,
      title={{ZeRO}: Memory Optimizations Toward Training Trillion Parameter Models}, 
      author={Samyam Rajbhandari and Jeff Rasley and Olatunji Ruwase and Yuxiong He},
      year={2020},
      booktitle={The International Conference for High Performance Computing, Networking, Storage and Analysis},
      url={https://arxiv.org/abs/1910.02054}, 
}

@inproceedings{
gudibande2024the,
title={The False Promise of Imitating Proprietary Language Models},
author={Arnav Gudibande and Eric Wallace and Charlie Victor Snell and Xinyang Geng and Hao Liu and Pieter Abbeel and Sergey Levine and Dawn Song},
booktitle={The Twelfth International Conference on Learning Representations (ICLR)},
year={2024},
url={https://openreview.net/forum?id=Kz3yckpCN5}
}

@misc{alpaca,
  author = {Rohan Taori and Ishaan Gulrajani and Tianyi Zhang and Yann Dubois and Xuechen Li and Carlos Guestrin and Percy Liang and Tatsunori B. Hashimoto },
  title = {Stanford Alpaca: An Instruction-following LLaMA model},
  year = {2023},
  publisher = {GitHub},
  journal = {GitHub repository},
  howpublished = {\url{https://github.com/tatsu-lab/stanford_alpaca}},
}

@inproceedings{huang-etal-2024-chat,
    title = "Chat Vector: A Simple Approach to Equip {LLM}s with Instruction Following and Model Alignment in New Languages",
    author = "Huang, Shih-Cheng  and
      Li, Pin-Zu  and
      Hsu, Yu-chi  and
      Chen, Kuang-Ming  and
      Lin, Yu Tung  and
      Hsiao, Shih-Kai  and
      Tsai, Richard  and
      Lee, Hung-yi",
    booktitle = "Proceedings of the 62nd Annual Meeting of the Association for Computational Linguistics (ACL)",
    month = aug,
    year = "2024",
    address = "Bangkok, Thailand",
    publisher = "Association for Computational Linguistics",
    url = "https://aclanthology.org/2024.acl-long.590/",
    doi = "10.18653/v1/2024.acl-long.590",
    pages = "10943--10959",
}

@misc{olmo2025olmo3,
      title={Olmo 3}, 
      author={Team Olmo : 
      Allyson Ettinger 
      and Amanda Bertsch 
      and Bailey Kuehl 
      and David Graham 
      and David Heineman 
      and Dirk Groeneveld 
      and Faeze Brahman 
      and Finbarr Timbers 
      and Hamish Ivison 
      and Jacob Morrison 
      and Jake Poznanski 
      and Kyle Lo 
      and Luca Soldaini 
      and Matt Jordan 
      and Mayee Chen 
      and others},
      year={2025},
      eprint={2512.13961},
      archivePrefix={arXiv},
      primaryClass={cs.CL},
      url={https://arxiv.org/abs/2512.13961}, 
      howpublished={arXiv:2512.13961},
}

@inproceedings{wang2022supernaturalinstructionsgeneralizationdeclarativeinstructions,
    title = "Super-{N}atural{I}nstructions: Generalization via Declarative Instructions on 1600+ {NLP} Tasks",
    author = "Wang, Yizhong  and
      Mishra, Swaroop  and
      Alipoormolabashi, Pegah  and
      Kordi, Yeganeh  and
      Mirzaei, Amirreza  and
      Naik, Atharva  and
      Ashok, Arjun  and
      Dhanasekaran, Arut Selvan  and
      Arunkumar, Anjana  and
      Stap, David  and
      Pathak, Eshaan  and
      Karamanolakis, Giannis  and
      Lai, Haizhi  and
      Purohit, Ishan  and
      Mondal, Ishani  and
      others",
    booktitle = "Proceedings of the 2022 Conference on Empirical Methods in Natural Language Processing (EMNLP)",
    year = "2022",
    url = "https://aclanthology.org/2022.emnlp-main.340/",
}

@inproceedings{wang-etal-2023-self-instruct,
    title = "Self-Instruct: Aligning Language Models with Self-Generated Instructions",
    author = "Wang, Yizhong  and
      Kordi, Yeganeh  and
      Mishra, Swaroop  and
      Liu, Alisa  and
      Smith, Noah A.  and
      Khashabi, Daniel  and
      Hajishirzi, Hannaneh",
    booktitle = "Proceedings of the 61st Annual Meeting of the Association for Computational Linguistics (ACL)",
    year = "2023",
    url = "https://aclanthology.org/2023.acl-long.754/",
}

@misc{openai2025gptoss120bgptoss20bmodel,
      title={gpt-oss-120b \& gpt-oss-20b Model Card}, 
      author={OpenAI},
      year={2025},
      eprint={2508.10925},
      archivePrefix={arXiv},
      primaryClass={cs.CL},
      url={https://arxiv.org/abs/2508.10925}, 
}

@misc{alpaca_eval,
  author = {Xuechen Li and Tianyi Zhang and Yann Dubois and Rohan Taori and Ishaan Gulrajani and Carlos Guestrin and Percy Liang and Tatsunori B. Hashimoto },
  title = {AlpacaEval: An Automatic Evaluator of Instruction-following Models},
  year = {2023},
  month = {5},
  publisher = {GitHub},
  journal = {GitHub repository},
  howpublished = {\url{https://github.com/tatsu-lab/alpaca_eval}}
}

@misc{gemmateam2025gemma3technicalreport,
      title={Gemma 3 Technical Report}, 
      author={Team Gemma 
      and Aishwarya Kamath 
      and Johan Ferret 
      and Shreya Pathak 
      and Nino Vieillard 
      and Ramona Merhej 
      and Sarah Perrin 
      and Tatiana Matejovicova 
      and Alexandre Ramé 
      and Morgane Rivière 
      and Louis Rouillard 
      and Thomas Mesnard 
      and Geoffrey Cideron 
      and Jean-bastien Grill 
      and Sabela Ramos 
      and Edouard Yvinec 
      and others},
      year={2025},
      eprint={2503.19786},
      archivePrefix={arXiv},
      primaryClass={cs.CL},
      url={https://arxiv.org/abs/2503.19786}, 
}

@misc{grattafiori2024llama3herdmodels,
      title={The Llama 3 Herd of Models}, 
      author={Aaron Grattafiori 
      and Abhimanyu Dubey 
      and Abhinav Jauhri 
      and Abhinav Pandey 
      and Abhishek Kadian 
      and Ahmad Al-Dahle 
      and Aiesha Letman 
      and Akhil Mathur 
      and Alan Schelten 
      and Alex Vaughan 
      and Amy Yang 
      and Angela Fan 
      and Anirudh Goyal 
      and Anthony Hartshorn 
      and Aobo Yang 
      and others},
      year={2024},
      eprint={2407.21783},
      archivePrefix={arXiv},
      primaryClass={cs.AI},
      url={https://arxiv.org/abs/2407.21783}, 
}

@inproceedings{
rafailov2023direct,
title={Direct Preference Optimization: Your Language Model is Secretly a Reward Model},
author={Rafael Rafailov and Archit Sharma and Eric Mitchell and Christopher D Manning and Stefano Ermon and Chelsea Finn},
booktitle={Thirty-seventh Conference on Neural Information Processing Systems (NeurIPS)},
year={2023},
url={https://openreview.net/forum?id=HPuSIXJaa9}
}

@inproceedings{
    xu2025magpie,
    title={Magpie: Alignment Data Synthesis from Scratch by Prompting Aligned   {LLM}s with Nothing},
    author={Zhangchen Xu and Fengqing Jiang and Luyao Niu and Yuntian Deng and  Radha Poovendran and Yejin Choi and Bill Yuchen Lin},
    booktitle={The Thirteenth International Conference on Learning  Representations},
    year={2025},
    url={https://openreview.net/forum?id=Pnk7vMbznK}
}

@inproceedings{
    zheng2024lmsyschatm,
    title={{LMSYS}-Chat-1M: A Large-Scale Real-World {LLM} Conversation Dataset},
    author={Lianmin Zheng and Wei-Lin Chiang and Ying Sheng and Tianle Li and Siyuan Zhuang and Zhanghao Wu and Yonghao Zhuang and Zhuohan Li and Zi Lin and Eric Xing and Joseph E. Gonzalez and Ion Stoica and Hao Zhang},
    booktitle={The Twelfth International Conference on Learning Representations (ICLR)},
    year={2024},
    url={https://openreview.net/forum?id=BOfDKxfwt0}
}

@misc{zhang2023modsmodelorienteddata,
      title={MoDS: Model-oriented Data Selection for Instruction Tuning}, 
      author={Qianlong Du and Chengqing Zong and Jiajun Zhang},
      year={2023},
      eprint={2311.15653},
      archivePrefix={arXiv},
      primaryClass={cs.CL},
      url={https://arxiv.org/abs/2311.15653}, 
}

@inproceedings{jiang-etal-2025-instruction,
    title = "Instruction-Tuning Data Synthesis from Scratch via Web Reconstruction",
    author = "Jiang, Yuxin  and
      Wang, Yufei  and
      Wu, Chuhan  and
      Dai, Xinyi  and
      Xu, Yan  and
      Gan, Weinan  and
      Wang, Yasheng  and
      Jiang, Xin  and
      Shang, Lifeng  and
      Tang, Ruiming  and
      Wang, Wei",
    editor = "Che, Wanxiang  and
      Nabende, Joyce  and
      Shutova, Ekaterina  and
      Pilehvar, Mohammad Taher",
    booktitle = "Findings of the Association for Computational Linguistics: ACL 2025",
    month = jul,
    year = "2025",
    publisher = "Association for Computational Linguistics",
    url = "https://aclanthology.org/2025.findings-acl.343/",
    doi = "10.18653/v1/2025.findings-acl.343",
    pages = "6603--6618",
    ISBN = "979-8-89176-256-5",
}

@inproceedings{
    yue2024mammoth,
    title={{MA}mmo{TH}2: Scaling Instructions from the Web},
    author={Xiang Yue and Tianyu Zheng and Ge Zhang and Wenhu Chen},
    booktitle={The Thirty-eighth Annual Conference on Neural Information Processing Systems (NeurIPS)},
    year={2024},
    url={https://openreview.net/forum?id=yVu5dnPlqA}
}

@inproceedings{
    li2024selfalignment,
    title={Self-Alignment with Instruction Backtranslation},
    author={Xian Li and Ping Yu and Chunting Zhou and Timo Schick and Omer Levy and Luke Zettlemoyer and Jason E Weston and Mike Lewis},
    booktitle={The Twelfth International Conference on Learning Representations (ICLR)},
    year={2024},
    url={https://openreview.net/forum?id=1oijHJBRsT}
}

@inproceedings{nguyen-etal-2024-better,
    title = "Better Alignment with Instruction Back-and-Forth Translation",
    author = "Nguyen, Thao  and
      Li, Jeffrey  and
      Oh, Sewoong  and
      Schmidt, Ludwig  and
      Weston, Jason E  and
      Zettlemoyer, Luke  and
      Li, Xian",
    editor = "Al-Onaizan, Yaser  and
      Bansal, Mohit  and
      Chen, Yun-Nung",
    booktitle = "Findings of the Association for Computational Linguistics: EMNLP 2024",
    month = nov,
    year = "2024",
    publisher = "Association for Computational Linguistics",
    url = "https://aclanthology.org/2024.findings-emnlp.777/",
    doi = "10.18653/v1/2024.findings-emnlp.777",
    pages = "13289--13308",
}

@inproceedings{wang-etal-2024-codeclm,
    title = "{C}odec{LM}: Aligning Language Models with Tailored Synthetic Data",
    author = "Wang, Zifeng  and
      Li, Chun-Liang  and
      Perot, Vincent  and
      Le, Long  and
      Miao, Jin  and
      Zhang, Zizhao  and
      Lee, Chen-Yu  and
      Pfister, Tomas",
    editor = "Duh, Kevin  and
      Gomez, Helena  and
      Bethard, Steven",
    booktitle = "Findings of the Association for Computational Linguistics: NAACL 2024",
    month = jun,
    year = "2024",
    publisher = "Association for Computational Linguistics",
    url = "https://aclanthology.org/2024.findings-naacl.235/",
    doi = "10.18653/v1/2024.findings-naacl.235",
    pages = "3712--3729",
}

@inproceedings{zhao-etal-2024-tree,
    title = "Tree-Instruct: A Preliminary Study of the Intrinsic Relationship between Complexity and Alignment",
    author = "Zhao, Yingxiu  and
      Yu, Bowen  and
      Hui, Binyuan  and
      Yu, Haiyang  and
      Li, Minghao  and
      Huang, Fei  and
      Zhang, Nevin L.  and
      Li, Yongbin",
    editor = "Calzolari, Nicoletta  and
      Kan, Min-Yen  and
      Hoste, Veronique  and
      Lenci, Alessandro  and
      Sakti, Sakriani  and
      Xue, Nianwen",
    booktitle = "Proceedings of the 2024 Joint International Conference on Computational Linguistics, Language Resources and Evaluation (LREC-COLING 2024)",
    month = may,
    year = "2024",
    address = "Torino, Italia",
    publisher = "ELRA and ICCL",
    url = "https://aclanthology.org/2024.lrec-main.1460/",
    pages = "16776--16789"
}

@misc{ivison2023camelschangingclimateenhancing,
      title={Camels in a Changing Climate: Enhancing LM Adaptation with Tulu 2}, 
      author={Hamish Ivison and Yizhong Wang and Valentina Pyatkin and Nathan Lambert and Matthew Peters and Pradeep Dasigi and Joel Jang and David Wadden and Noah A. Smith and Iz Beltagy and Hannaneh Hajishirzi},
      year={2023},
      eprint={2311.10702},
      archivePrefix={arXiv},
      primaryClass={cs.CL},
      url={https://arxiv.org/abs/2311.10702}, 
}

@inproceedings{
    liu2024what,
    title={What Makes Good Data for Alignment? A Comprehensive Study of Automatic Data Selection in Instruction Tuning},
    author={Wei Liu and Weihao Zeng and Keqing He and Yong Jiang and Junxian He},
    booktitle={The Twelfth International Conference on Learning Representations (ICLR)},
    year={2024},
    url={https://openreview.net/forum?id=BTKAeLqLMw}
}

@inproceedings{li-etal-2024-quantity,
    title = "From Quantity to Quality: Boosting {LLM} Performance with Self-Guided Data Selection for Instruction Tuning",
    author = "Li, Ming  and
      Zhang, Yong  and
      Li, Zhitao  and
      Chen, Jiuhai  and
      Chen, Lichang  and
      Cheng, Ning  and
      Wang, Jianzong  and
      Zhou, Tianyi  and
      Xiao, Jing",
    editor = "Duh, Kevin  and
      Gomez, Helena  and
      Bethard, Steven",
    booktitle = "Proceedings of the 2024 Conference of the North American Chapter of the Association for Computational Linguistics: Human Language Technologies (NAACL)",
    month = jun,
    year = "2024",
    publisher = "Association for Computational Linguistics",
    url = "https://aclanthology.org/2024.naacl-long.421/",
    doi = "10.18653/v1/2024.naacl-long.421",
    pages = "7602--7635",
}

@misc{
    obrien2024contrastive,
    title={Contrastive Decoding Improves Reasoning in Large Language Models},
    author={Sean O'Brien and Mike Lewis},
    year={2024},
    url={https://openreview.net/forum?id=SzV37yefM4}
}

@inproceedings{chang-etal-2024-explaining,
    title = "Explaining and Improving Contrastive Decoding by Extrapolating the Probabilities of a Huge and Hypothetical {LM}",
    author = "Chang, Haw-Shiuan  and
      Peng, Nanyun  and
      Bansal, Mohit  and
      Ramakrishna, Anil  and
      Chung, Tagyoung",
    editor = "Al-Onaizan, Yaser  and
      Bansal, Mohit  and
      Chen, Yun-Nung",
    booktitle = "Proceedings of the 2024 Conference on Empirical Methods in Natural Language Processing (EMNLP)",
    month = nov,
    year = "2024",
    publisher = "Association for Computational Linguistics",
    url = "https://aclanthology.org/2024.emnlp-main.484/",
    doi = "10.18653/v1/2024.emnlp-main.484",
    pages = "8503--8526",
}

@inproceedings{liu-etal-2021-dexperts,
    title = "{DE}xperts: Decoding-Time Controlled Text Generation with Experts and Anti-Experts",
    author = "Liu, Alisa  and
      Sap, Maarten  and
      Lu, Ximing  and
      Swayamdipta, Swabha  and
      Bhagavatula, Chandra  and
      Smith, Noah A.  and
      Choi, Yejin",
    editor = "Zong, Chengqing  and
      Xia, Fei  and
      Li, Wenjie  and
      Navigli, Roberto",
    booktitle = "Proceedings of the 59th Annual Meeting of the Association for Computational Linguistics and the 11th International Joint Conference on Natural Language Processing (Volume 1: Long Papers)",
    month = aug,
    year = "2021",
    address = "Online",
    publisher = "Association for Computational Linguistics",
    url = "https://aclanthology.org/2021.acl-long.522/",
    doi = "10.18653/v1/2021.acl-long.522",
    pages = "6691--6706"
}

@inproceedings{
    chuang2024dola,
    title={DoLa: Decoding by Contrasting Layers Improves Factuality in Large Language Models},
    author={Yung-Sung Chuang and Yujia Xie and Hongyin Luo and Yoon Kim and James R. Glass and Pengcheng He},
    booktitle={The Twelfth International Conference on Learning Representations (ICLR)},
    year={2024},
    url={https://openreview.net/forum?id=Th6NyL07na}
}

@inproceedings{zhang-etal-2025-alleviating,
    title = "Alleviating Hallucinations of Large Language Models through Induced Hallucinations",
    author = "Zhang, Yue  and
      Cui, Leyang  and
      Bi, Wei  and
      Shi, Shuming",
    editor = "Chiruzzo, Luis  and
      Ritter, Alan  and
      Wang, Lu",
    booktitle = "Findings of the Association for Computational Linguistics: NAACL 2025",
    month = apr,
    year = "2025",
    publisher = "Association for Computational Linguistics",
    url = "https://aclanthology.org/2025.findings-naacl.459/",
    doi = "10.18653/v1/2025.findings-naacl.459",
    pages = "8233--8247",
    ISBN = "979-8-89176-195-7",
}

@inproceedings{shi-etal-2024-trusting,
    title = "Trusting Your Evidence: Hallucinate Less with Context-aware Decoding",
    author = "Shi, Weijia  and
      Han, Xiaochuang  and
      Lewis, Mike  and
      Tsvetkov, Yulia  and
      Zettlemoyer, Luke  and
      Yih, Wen-tau",
    editor = "Duh, Kevin  and
      Gomez, Helena  and
      Bethard, Steven",
    booktitle = "Proceedings of the 2024 Conference of the North American Chapter of the Association for Computational Linguistics: Human Language Technologies (NAACL)",
    month = jun,
    year = "2024",
    publisher = "Association for Computational Linguistics",
    url = "https://aclanthology.org/2024.naacl-short.69/",
    doi = "10.18653/v1/2024.naacl-short.69",
    pages = "783--791",
}

@inproceedings{
    sanchez2024stay,
    title={Stay on Topic with Classifier-Free Guidance},
    author={Guillaume Sanchez and Alexander Spangher and Honglu Fan and Elad Levi and Stella Biderman},
    booktitle={Forty-first International Conference on Machine Learning (ICML)},
    year={2024},
    url={https://openreview.net/forum?id=RiM3cl9MdK}
}

@inproceedings{
    lu2024opendomain,
    title={Open-Domain Text Evaluation via Contrastive Distribution Methods},
    author={Sidi Lu and Hongyi Liu and Asli Celikyilmaz and Tianlu Wang and Nanyun Peng},
    booktitle={Forty-first International Conference on Machine Learning (ICML)},
    year={2024},
    url={https://openreview.net/forum?id=9HdQr68Zyl}
}

@inproceedings{
    jiang2026contrastive,
    title={Contrastive Weak-to-Strong Generalization},
    author={Houcheng Jiang and Junfeng Fang and Jiaxin Wu and Tianyu Zhang and Chen Gao and Xiang Wang and Xiangnan He and Yang Deng},
    booktitle={Forty-third International Conference on Machine Learning (ICML)},
    year={2026},
    url={https://openreview.net/forum?id=maOM1OWNZ6}
}

@inproceedings{
    lu2025unidetox,
    title={UniDetox: Universal Detoxification of Large Language Models via Dataset Distillation},
    author={Huimin LU and Masaru Isonuma and Junichiro Mori and Ichiro Sakata},
    booktitle={The Thirteenth International Conference on Learning Representations (ICLR)},
    year={2025},
    url={https://openreview.net/forum?id=eLLBILFRsA}
}

@inproceedings{
    isonuma2025whats,
    title={What's New in My Data? Novelty Exploration via Contrastive Generation},
    author={Masaru Isonuma and Ivan Titov},
    booktitle={The Thirteenth International Conference on Learning Representations (ICLR)},
    year={2025},
    url={https://openreview.net/forum?id=IZDiRbVSVN}
}

@inproceedings{
    liu2024tuning,
    title={Tuning Language Models by Proxy},
    author={Alisa Liu and Xiaochuang Han and Yizhong Wang and Yulia Tsvetkov and Yejin Choi and Noah A. Smith},
    booktitle={First Conference on Language Modeling (COLM)},
    year={2024},
    url={https://openreview.net/forum?id=dribhnhm1i}
}

@inproceedings{
    gao2024linear,
    title={Linear Alignment: A Closed-form Solution for Aligning Human Preferences without Tuning and Feedback},
    author={Songyang Gao and Qiming Ge and Wei Shen and Shihan Dou and Junjie Ye and Xiao Wang and Rui Zheng and Yicheng Zou and Zhi Chen and Hang Yan and Qi Zhang and Dahua Lin},
    booktitle={Forty-first International Conference on Machine Learning (ICML)},
    year={2024},
    url={https://openreview.net/forum?id=Y4wxCICbD0}
}

@inproceedings{hu-etal-2025-nile,
    title = "{NILE}: Internal Consistency Alignment in Large Language Models",
    author = "Hu, Minda  and
      Zhang, Qiyuan  and
      Wang, Yufei  and
      He, Bowei  and
      Wang, Hongru  and
      Zhou, Jingyan  and
      Li, Liangyou  and
      Wang, Yasheng  and
      Ma, Chen  and
      King, Irwin",
    editor = "Christodoulopoulos, Christos  and
      Chakraborty, Tanmoy  and
      Rose, Carolyn  and
      Peng, Violet",
    booktitle = "Proceedings of the 2025 Conference on Empirical Methods in Natural Language Processing (EMNLP)",
    month = nov,
    year = "2025",
    publisher = "Association for Computational Linguistics",
    url = "https://aclanthology.org/2025.emnlp-main.412/",
    doi = "10.18653/v1/2025.emnlp-main.412",
    pages = "8129--8147",
    ISBN = "979-8-89176-332-6",
}

@inproceedings{li-etal-2025-mosaic,
    title = "Mosaic-{IT}: Cost-Free Compositional Data Synthesis for Instruction Tuning",
    author = "Li, Ming  and
      Chen, Pei  and
      Wang, Chenguang  and
      Zhao, Hongyu  and
      Liang, Yijun  and
      Hou, YuPeng  and
      Liu, Fuxiao  and
      Zhou, Tianyi",
    editor = "Che, Wanxiang  and
      Nabende, Joyce  and
      Shutova, Ekaterina  and
      Pilehvar, Mohammad Taher",
    booktitle = "Findings of the Association for Computational Linguistics: ACL 2025",
    month = jul,
    year = "2025",
    address = "Vienna, Austria",
    publisher = "Association for Computational Linguistics",
    url = "https://aclanthology.org/2025.findings-acl.1297/",
    doi = "10.18653/v1/2025.findings-acl.1297",
    pages = "25287--25318",
    ISBN = "979-8-89176-256-5"
}

@inproceedings{
    lambert2025tulu,
    title={Tulu 3: Pushing Frontiers in Open Language Model Post-Training},
    author={Nathan Lambert and Jacob Morrison and Valentina Pyatkin and Shengyi Huang and Hamish Ivison and Faeze Brahman and Lester James Validad Miranda and Alisa Liu and Nouha Dziri and Xinxi Lyu and Yuling Gu and Saumya Malik and Victoria Graf and Jena D. Hwang and Jiangjiang Yang and Ronan Le Bras and Oyvind Tafjord and Christopher Wilhelm and Luca Soldaini and Noah A. Smith and Yizhong Wang and Pradeep Dasigi and Hannaneh Hajishirzi},
    booktitle={Second Conference on Language Modeling (COLM)},
    year={2025},
    url={https://openreview.net/forum?id=i1uGbfHHpH}
}

@article{DBLP:journals/corr/abs-2501-12948,
  publtype={informal},
  author={DeepSeek-AI 
  and Daya Guo 
  and Dejian Yang 
  and Haowei Zhang 
  and Junxiao Song 
  and Ruoyu Zhang 
  and Runxin Xu 
  and Qihao Zhu 
  and Shirong Ma 
  and Peiyi Wang 
  and Xiao Bi 
  and Xiaokang Zhang 
  and Xingkai Yu 
  and Yu Wu and Z. F. Wu 
  and Zhibin Gou 
  and Zhihong Shao 
  and others},
  title={DeepSeek-R1: Incentivizing Reasoning Capability in LLMs via Reinforcement Learning},
  year={2025},
  month={January},
  cdate={1735689600000},
  journal={CoRR},
  volume={abs/2501.12948},
  url={https://doi.org/10.48550/arXiv.2501.12948}
}

@inproceedings{
    yue2025does,
    title={Does Reinforcement Learning Really Incentivize Reasoning Capacity in {LLM}s Beyond the Base Model?},
    author={Yang Yue and Zhiqi Chen and Rui Lu and Andrew Zhao and Zhaokai Wang and Yang Yue and Shiji Song and Gao Huang},
    booktitle={The Thirty-ninth Annual Conference on Neural Information Processing Systems (NeurIPS)},
    year={2025},
    url={https://openreview.net/forum?id=4OsgYD7em5}
}

@inproceedings{
    matsutani2026rl,
    title={{RL} Squeezes, {SFT} Expands: A Comparative Study of Reasoning {LLM}s},
    author={Kohsei Matsutani and Shota Takashiro and Gouki Minegishi and Takeshi Kojima and Yusuke Iwasawa and Yutaka Matsuo},
    booktitle={The Fourteenth International Conference on Learning Representations (ICLR)},
    year={2026},
    url={https://openreview.net/forum?id=N2lMNqJsBw}
}

@inproceedings{
    liu2025understanding,
    title={Understanding R1-Zero-Like Training: A Critical Perspective},
    author={Zichen Liu and Changyu Chen and Wenjun Li and Penghui Qi and Tianyu Pang and Chao Du and Wee Sun Lee and Min Lin},
    booktitle={Second Conference on Language Modeling (COLM)},
    year={2025},
    url={https://openreview.net/forum?id=5PAF7PAY2Y}
}

@inproceedings{
    zhao2025echo,
    title={Echo Chamber: {RL} Post-training Amplifies Behaviors Learned in Pretraining},
    author={Rosie Zhao and Alexandru Meterez and Sham M. Kakade and Cengiz Pehlevan and Samy Jelassi and Eran Malach},
    booktitle={Second Conference on Language Modeling (COLM)},
    year={2025},
    url={https://openreview.net/forum?id=dp4KWuSDzj}
}

@misc{openai2024gpt4technicalreport,
      title={GPT-4 Technical Report}, 
      author={OpenAI 
      and Josh Achiam 
      and Steven Adler 
      and Sandhini Agarwal 
      and Lama Ahmad 
      and Ilge Akkaya 
      and Florencia Leoni Aleman 
      and Diogo Almeida 
      and Janko Altenschmidt 
      and Sam Altman 
      and Shyamal Anadkat 
      and Red Avila 
      and Igor Babuschkin 
      and Suchir Balaji 
      and Valerie Balcom 
      and Paul Baltescu 
      and others
      },
      year={2024},
      eprint={2303.08774},
      archivePrefix={arXiv},
      primaryClass={cs.CL},
      url={https://arxiv.org/abs/2303.08774}, 
}

@misc{singh2026openaigpt5card,
      title={OpenAI GPT-5 System Card}, 
      author={OpenAI 
      and Aaditya Singh 
      and Adam Fry 
      and Adam Perelman 
      and Adam Tart 
      and Adi Ganesh 
      and Ahmed El-Kishky 
      and Aidan McLaughlin 
      and Aiden Low 
      and AJ Ostrow 
      and Akhila Ananthram 
      and Akshay Nathan 
      and Alan Luo 
      and Alec Helyar 
      and Aleksander Madry 
      and Aleksandr Efremov 
      and others},
      year={2026},
      eprint={2601.03267},
      archivePrefix={arXiv},
      primaryClass={cs.CL},
      url={https://arxiv.org/abs/2601.03267}, 
}
\bibliographystyle{colm2026_conference}

\newpage
\appendix

\section{Generation details}
\label{generation_param}

\subsection{Hyperparameters for response generation}
\label{subsec:generation_params}
To ensure a fair comparison, the hyperparameters for response generation were kept consistent across all experiments. We used a temperature of 1.0, a Top-$p$ value of 1.0, and a maximum generation length of 4,096 tokens.

\subsection{Best-of-N selection and evaluation prompt}
\label{appendix:best_of_n}
\begin{figure}
    \centering
    \includegraphics[width=0.5\linewidth]{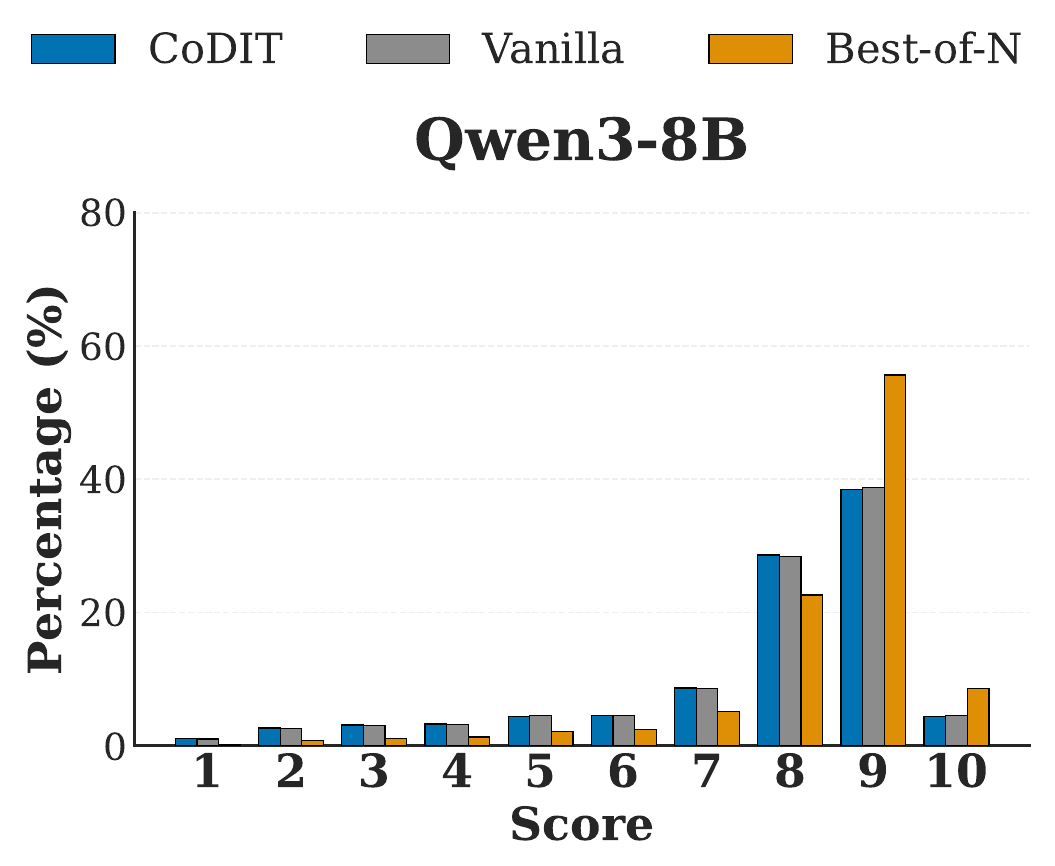}
    \caption{Score distribution of synthesized responses using Qwen3-8B. Consistent with the results in Section~\ref{sec:score_distribution}, \methodname~maintains a text quality comparable to the Vanilla teacher model, indicating that the method's effectiveness does not rely on enhancing general text fluency.}
    \label{fig:score_distribution_qwen3}
\end{figure}

For the Best-of-N selection process, we utilized gpt-oss-120b~\citep{openai2025gptoss120bgptoss20bmodel} as a judge to evaluate the candidate responses. The evaluation was conducted using a prompt adapted from the WildBench evaluation framework~\citep{lin2025wildbench}. 
Each response was rated on a scale from 1 to 10. From the 5 generated candidates, we selected the response that received the highest score. When multiple responses achieved the same maximum score, we consistently selected the first one.
The scoring results for the data synthesized using Qwen3-8B~\citep{yang2025qwen3technicalreport} are also provided in Figure~\ref{fig:score_distribution_qwen3}.

\begin{center}
\begin{tcolorbox}[
    breakable,
    title=Annotation Prompt Used in Best-of-N, 
]
\begin{lstlisting}[breaklines=true]
# Instruction 

You are an expert evaluator. Your task is to evaluate the quality of the responses generated by AI models. 
We will provide you with the user query and an AI-generated response.
You should first read the user query carefully for analyzing the task, and then evaluate the quality of the response based on the rules provided below.

# Conversation between User and AI

## Current User Query
<|begin_of_query|>

{question}

<|end_of_query|>

## AI Response
<|begin_of_response|>

{answer}

<|end_of_response|>
 

# Evaluation   

## Rules 

You should evaluate the above response based on your analysis of the user queries.
You should first write down your analysis, and then provide your assessment.
The scores are in the range of 1~10, where 1 means the response is very poor and 10 means the response is perfect.
Here are more detailed criteria for the scores:

- Score 1~2: The response is very poor and does not make sense at all.
- Score 3~4: The response is poor and does not help user solve the problem in a meaningful way.
- Score 5~6: The response is fair but has some issues (e.g., factual errors, hallucinations, missing key information).
- Score 7~8: The response is good enough but could be improved in some ways.
- Score 9~10: The response is perfect and provides helpful information that can help user solve the problem.

## Output Format 
First, please output your analysis for the model response, and then summarize your assessment to two aspects: \"strengths\" and \"weaknesses\"; Finally, please write down your rating for the assessment.

Please provide your evaluation results in the following json format by filling in the placeholders in []:
```
{
    \"strengths\": \"[analysis for the strengths of the response]\",
    \"weaknesses\": \"[analysis for the weaknesses of the response]\",
    \"score\": \"[1~10]\"
}
```
\end{lstlisting}
\label{fig: evaluation prompt best-of-n}
\end{tcolorbox}

\end{center}

\section{Training details}
\label{appendix:generation_hyp}

\begin{table}[t]
\centering
\begin{tabular}{ll}
\toprule
\textbf{Hyperparameter} & \textbf{Value} \\
\midrule
Optimizer & AdamW ($\beta_1 = 0.90, \beta_2 = 0.95$) \\
Learning rate scheduler & Cosine with 0.1 warmup ratio \\
Peak learning rate & $2.5 \times 10^{-5}$ ($1.0 \times 10^{-5}$ for gemma-3-4b-pt) \\
Minimum learning rate & $2.5 \times 10^{-6}$ ($1.0 \times 10^{-6}$ for gemma-3-4b-pt) \\
Training epochs & 2 \\
Effective batch size & 512 \\
\bottomrule
\end{tabular}
\caption{Hyperparameters for Supervised Fine-Tuning}
\label{tab:hyperparameters}
\end{table}

All training was conducted on four NVIDIA H100 SXM5 GPUs, utilizing DeepSpeed ZeRO \citep{rajbhandari2020zeromemoryoptimizationstraining} for memory optimization. Unless otherwise specified, hyperparameters followed the settings established in prior work \citep{ma2025building}; the specific values used in our experiments are summarized in Table \ref{tab:hyperparameters}. However, since performance degradation was observed for gemma-3-4b-pt~\citep{gemmateam2025gemma3technicalreport} under these default settings, we conducted a learning rate sweep using MT-Bench~\citep{zheng2023judging} as the metric. Consequently, for this specific model, the peak and minimum learning rates were adjusted to $1.0 \times 10^{-5}$ and $1.0 \times 10^{-6}$, respectively.

For data preprocessing, several constraints were applied to manage memory limitations. For the gemma-3-4b-pt model trained on the Qwen3-30B-A3B-distilled dataset, responses exceeding 100,000 characters (3 instances) were truncated to that limit to avoid Out-of-Memory (OOM) errors. Regarding WildChat \citep{zhao2024wildchat}, Magpie-Pro \citep{xu2025magpie}, WebR-Basic, and WebR-Pro \citep{jiang-etal-2025-instruction}, we excluded instances where the instruction exceeded 4,096 characters or the response exceeded 8,192 characters. Additionally, for WildChat, only the first turn of each conversation was utilized.

\section{\texorpdfstring{Ablation study on $\alpha$}{Ablation Study on alpha}}
\label{appendix:alpha_ablation}

\subsection{\texorpdfstring{$\alpha$ tuning}{alpha Tuning}}
\label{appendix:alpha_tuning}

\begin{figure}
    \centering
    \includegraphics[width=0.5\linewidth]{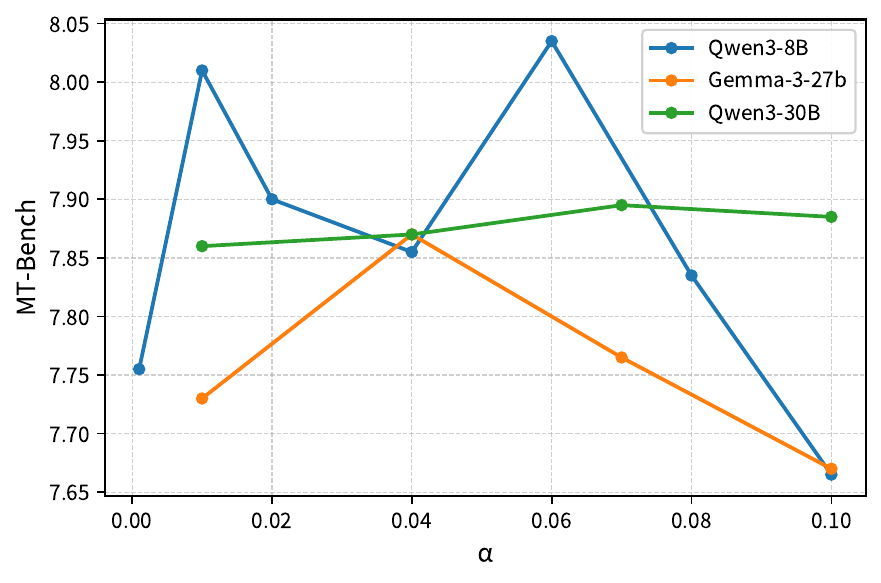}
    \caption{Hyperparameter tuning for $\alpha$ based on MT-Bench scores. The average performance of student models (Qwen3-8B-Base and Llama-3.1-8B) is plotted against different $\alpha$ values for each teacher model. The optimal $\alpha$ is selected where the MT-Bench score is maximized.}
    \label{fig:cd-alpha-ablation}
\end{figure}

To determine the optimal value for the hyperparameter $\alpha$ in Equation \eqref{eq:contrastive_decoding}, we utilized MT-Bench~\citep{zheng2023judging} as our validation benchmark. For this tuning process, we used the same set of instructions as described in Section \ref{sec:experimental_setup}. We constructed multiple variants of synthetic datasets by applying \methodname~across a specific grid of $\alpha$ values: $\{0.001, 0.01, 0.02, 0.04, 0.06, 0.08, 0.1\}$ for Qwen3-8B~\citep{yang2025qwen3technicalreport}, and $\{0.01, 0.04, 0.07, 0.1\}$ for Qwen3-30B-A3B~\citep{yang2025qwen3technicalreport} and gemma-3-27b-it~\citep{gemmateam2025gemma3technicalreport}. 

For each dataset, we performed instruction tuning on two pre-trained models: Qwen3-8B-Base~\citep{yang2025qwen3technicalreport} and Llama-3.1-8B~\citep{grattafiori2024llama3herdmodels}. We then selected the $\alpha$ value that maximized the average MT-Bench score across both models. As shown in Figure \ref{fig:cd-alpha-ablation}, which plots the relationship between the choice of $\alpha$ and the average performance, our experimental results identified the optimal values as $\alpha = 0.06$ for Qwen3-8B, $\alpha = 0.04$ for gemma-3-27b-it, and $\alpha = 0.07$ for Qwen3-30B-A3B. Based on these findings, we adopted these specific values for each teacher model in all subsequent experiments in this study.
Furthermore, for response generation with Llama-3.2-1B, Llama-3.2-3B~\citep{grattafiori2024llama3herdmodels}, and Llama-3.1-8B, we used $\alpha = 0.06$, consistent with the value determined for Qwen3-8B.

\subsection{Robustness to hyperparameter \texorpdfstring{$\alpha$}{alpha}}

\begin{figure}
    \centering
    \includegraphics[width=1\linewidth]{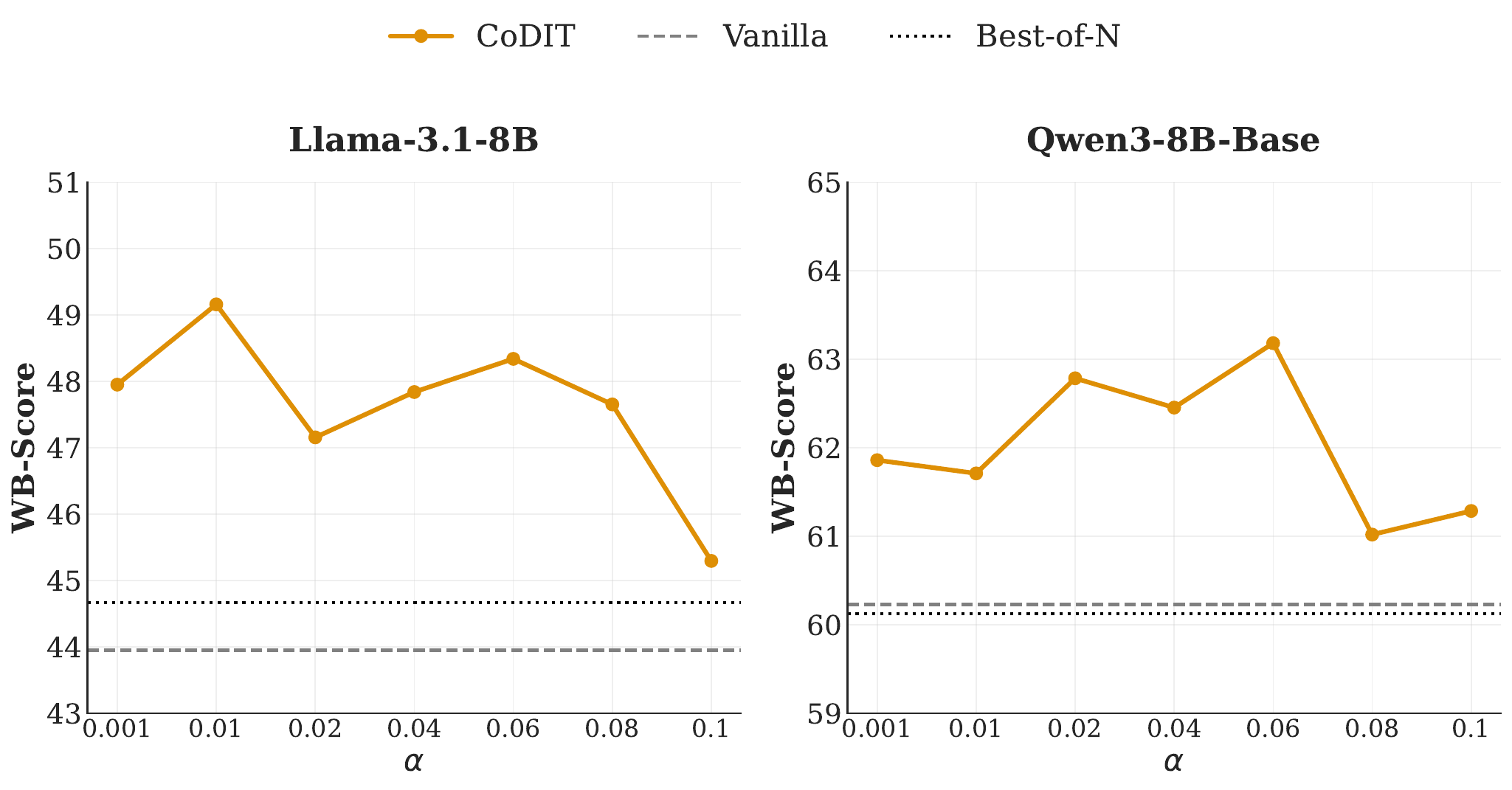}
    \caption{Robustness of \methodname~across various $\alpha$ settings. Performance on WildBench remains stable and superior to the baselines regardless of the specific choice of $\alpha$, indicating that the gains are driven by the core \methodname~mechanism rather than sensitive hyperparameter tuning.}
    \label{fig:alpha-ablation-wb}
\end{figure}

A potential concern is whether the performance gains of \methodname~are overly dependent on the specific tuning of $\alpha$ performed in Section \ref{appendix:alpha_tuning}. To address this, we evaluated the sensitivity of \methodname~to the choice of $\alpha$ using WildBench.
Using the same set of instructions as described in Section \ref{sec:experimental_setup} and Qwen3-8B as the teacher, we compared models trained on datasets generated with $\alpha \in \{0.001, 0.01, 0.02, 0.04, 0.06, 0.08, 0.1\}$. The results, as shown in Figure \ref{fig:alpha-ablation-wb}, demonstrate that both student models achieved consistently higher scores across all tested values of $\alpha$ compared to the baselines. This indicates that the improvement is primarily driven by the core mechanism of \methodname~rather than exhaustive hyperparameter optimization, confirming the robustness of our approach.

\section{Evaluation prompts for WildBench}
\label{appendix:wildbench-prompt}

\begin{table*}[t]
\small
\centering
\setlength{\tabcolsep}{4pt}
\begin{tabular}{l ccc ccc ccc}
\toprule
Teacher & \multicolumn{3}{c}{\textbf{Qwen3-8B}} & \multicolumn{3}{c}{\textbf{Qwen3-30B}} & \multicolumn{3}{c}{\textbf{Gemma-3-27B}} \\
\cmidrule(lr){1-1} \cmidrule(lr){2-4} \cmidrule(lr){5-7} \cmidrule(lr){8-10}
Student & Llama & Qwen & Gemma & Llama & Qwen & Gemma & Llama & Qwen & Gemma \\
\midrule
Vanilla & 43.45 & 60.12 & 30.21 & 43.01 & 57.55 & 31.37 & 47.40 & 59.63 & 34.08
\\
\rowcolor{gray!10} \methodname & \textbf{48.50} & \textbf{62.97} & \textbf{34.70} & \textbf{48.19} & \textbf{60.43} & \textbf{34.06} & \textbf{54.96} & \textbf{62.92} & \textbf{38.91} \\
\bottomrule
\end{tabular}
\caption{WB-Score on WildBench using the original official prompts. It can be observed that \methodname~consistently yields performance improvements compared to the direct use of teacher model outputs (Vanilla), even when evaluated with the official prompts.}
\label{tab:wb_official_results}
\end{table*}

Due to several grammatical and contextual errors identified in the official WildBench evaluation prompts, we applied corrections to ensure an accurate and reliable assessment in this study.
Furthermore, we report the evaluation results obtained with the original official prompts in Table \ref{tab:wb_official_results}. Our findings indicate that \methodname~maintains a performance lead over the Vanilla baseline across both the original and our corrected versions of the prompts.

\begin{center}
\begin{tcolorbox}[
    breakable,
    title=Evaluation Prompt for WildBench, 
]
\begin{lstlisting}[breaklines=true]
# Instruction 

You are an expert evaluator. Your task is to evaluate the quality of the responses generated by AI models. 
We will provide you with the user query and an AI-generated response.
You should first read the user query and the conversation history carefully for analyzing the task, and then evaluate the quality of the response based on the rules provided below.

# Conversation between User and AI

## History
<|begin_of_history|>

{$history}

<|end_of_history|> 

## Current User Query
<|begin_of_query|>

{$user_query}

<|end_of_query|>

## AI Response
<|begin_of_response|>

{$model_output}

<|end_of_response|>
 

# Evaluation   

## Checklist 

<|begin_of_checklist|>

{$checklist}

<|end_of_checklist|>

Please use this checklist to guide your evaluation, but do not limit your assessment to the checklist.

## Rules 

You should evaluate the above response based on your analysis of the user queries and the conversation history.
You should first write down your analysis and the checklist that you used for the evaluation, and then provide your assessment according to the checklist.
The scores are in the range of 1~10, where 1 means the response is very poor and 10 means the response is perfect.
Here are more detailed criteria for the scores:

- Score 1~2: The response is very poor and does not make sense at all.
- Score 3~4: The response is poor and does not help user solve the problem in a meaningful way.
- Score 5~6: The response is fair but has some issues (e.g., factual errors, hallucinations, missing key information).
- Score 7~8: The response is good enough but could be improved in some ways.
- Score 9~10: The response is perfect and provides helpful information that can help user solve the problem.

## Output Format 
First, please output your analysis for the model response, and then summarize your assessment to two aspects: \"strengths\" and \"weaknesses\"; Finally, please write down your rating for the assessment.

Please provide your evaluation results in the following json format by filling in the placeholders in []:
```
{
    "strengths": "[analysis for the strengths of the response]",
    "weaknesses": "[analysis for the weaknesses of the response]",
    "score": "[1~10]"
}
```
\end{lstlisting}
\end{tcolorbox}
\label{fig: generating mt prompt}
\end{center}

\section{Generalization to other instruction sources}
\label{app:webr_generalization}

\begin{table}[t]
\centering
\small
\begin{tabular}{l cc c}
\toprule
\multirow{2}{*}{\textbf{Method}} & \multicolumn{2}{c}{\textbf{AlpacaEval 2.0}} & \textbf{WildBench} \\
\cmidrule(lr){2-3} \cmidrule(lr){4-4}
 & LC & WR & WB-Score \\
\midrule
Vanilla & 37.65 & 46.99 & 42.61 \\
\rowcolor{gray!10}
\methodname & \bb{40.47} & \bb{51.86} & \bb{45.35} \\
\bottomrule
\end{tabular}
\caption{Effect of applying \methodname's response-generation procedure to the WebR-Pro instruction source, with Qwen3-8B as the teacher and Llama-3.1-8B as the student.}
\label{tab:webr_generalization}
\end{table}
Since \methodname~is a general-purpose response-generation method, it is not tied to any specific instruction source. It can therefore be readily applied to instruction sources other than LMSYS-Chat-1M~\citep{zheng2024lmsyschatm}, which we used in our main
experiments (Section~\ref{sec:main_results}), such as WebR-Pro~\citep{jiang-etal-2025-instruction}.

To empirically support this point, we applied CoDIT's response-generation procedure directly to the WebR-Pro instruction source and compared the resulting responses, using Qwen3-8B as the teacher model and Llama-3.1-8B as the student model. As shown in Table~\ref{tab:webr_generalization}, CoDIT-generated responses yield consistent gains across all metrics.

\section{Reliability of evaluation results}

\subsection{Robustness across teacher--student configurations}
\label{app:standard_errors}

\begin{table*}[t]
\small
\centering
\setlength{\tabcolsep}{4pt}
\begin{tabular}{l ccc ccc ccc}
\toprule
\textbf{Teacher} & \multicolumn{3}{c}{\textbf{Qwen3-8B}} & \multicolumn{3}{c}{\textbf{Qwen3-30B-A3B}} & \multicolumn{3}{c}{\textbf{Gemma-3-27B}} \\
\cmidrule(lr){1-1} \cmidrule(lr){2-4} \cmidrule(lr){5-7} \cmidrule(lr){8-10}
Student & Llama & Qwen & Gemma & Llama & Qwen & Gemma & Llama & Qwen & Gemma \\
\midrule

\multicolumn{10}{l}{\textbf{WildBench WB-Score}} \\
\midrule
Vanilla & 1.31 & 0.97 & 1.43 & 1.35 & 1.10 & 1.42 & 1.30 & 0.97 & 1.36 \\
Best-of-N & 1.28 & 0.99 & 1.41 & 1.23 & 0.97 & 1.44 & 1.22 & 0.98 & 1.40 \\
\rowcolor{gray!10} \methodname & 1.25 & 0.93 & 1.42 & 1.24 & 0.97 & 1.41 & 1.13 & 0.95 & 1.39 \\
\midrule

\multicolumn{10}{l}{\textbf{Alpaca Eval 2.0 Win Rates (\%)}} \\
\midrule
Vanilla & 1.57 & 1.53 & 1.53 & 1.58 & 1.55 & 1.52 & 1.45 & 1.42 & 1.55 \\
Best-of-N & 1.56 & 1.50 & 1.53 & 1.57 & 1.53 & 1.53 & 1.42 & 1.39 & 1.53 \\
\rowcolor{gray!10} \methodname & 1.57 & 1.46 & 1.57 & 1.57 & 1.54 & 1.56 & 1.35 & 1.33 & 1.51 \\
\midrule

\multicolumn{10}{l}{\textbf{Alpaca Eval 2.0 Length-Controlled Win Rates (\%)}} \\
\midrule
Vanilla & 0.61 & 0.64 & 0.62 & 0.67 & 0.69 & 0.64 & 0.65 & 0.65 & 0.64 \\
Best-of-N & 0.66 & 0.68 & 0.65 & 0.71 & 0.68 & 0.65 & 0.68 & 0.61 & 0.58 \\
\rowcolor{gray!10} \methodname & 0.58 & 0.53 & 0.55 & 0.58 & 0.56 & 0.58 & 0.62 & 0.62 & 0.56 \\
\midrule

\multicolumn{10}{l}{\textbf{MT-Bench}} \\
\midrule
Vanilla & 2.38 & 1.71 & 2.56 & 2.65 & 1.58 & 2.59 & 2.59 & 1.95 & 2.71 \\
Best-of-N & 2.57 & 1.75 & 2.51 & 2.53 & 1.84 & 2.52 & 2.75 & 2.03 & 2.70 \\
\rowcolor{gray!10} \methodname & 2.31 & 1.67 & 2.55 & 2.52 & 1.71 & 2.60 & 2.65 & 1.87 & 2.54 \\
\bottomrule
\end{tabular}
\caption{Standard errors for WildBench, MT-Bench, and AlpacaEval metrics}
\label{tab:standard_errors}
\end{table*}

Our experimental design evaluates \methodname~across nine distinct teacher--student configurations in Table~\ref{tab:main_results} and effectively serves as a repeated evaluation under varied conditions, providing stronger evidence of robustness than multi-seed runs of a single configuration. The consistent superiority of \methodname~on AlpacaEval 2.0 Win Rate in all nine configurations (9/9) and on WildBench WB-Score in eight of the nine configurations (8/9) across these diverse settings further supports the reliability of our findings. For further reference, we report the standard errors of the WildBench, AlpacaEval 2.0, and MT-Bench scores in Table~\ref{tab:standard_errors}.

\subsection{Evaluation with a more advanced judge model}
\label{app:gpt55_judge}
\begin{table*}[t]
\centering
\small
\begin{tabular}{l ccc ccc c}
\toprule
\multirow{2}{*}{\textbf{Method}} & \multicolumn{3}{c}{\textbf{gemma-3-27b-it}} & \multicolumn{3}{c}{\textbf{Qwen3-8B}} & \multicolumn{1}{c}{\textbf{Qwen3-30B-A3B}} \\
\cmidrule(lr){2-4} \cmidrule(lr){5-7} \cmidrule(lr){8-8}
 & Llama & Qwen & gemma & Llama & Qwen & gemma & Llama \\
\midrule
Vanilla & 4.07 & 11.38 & $-$13.15 & $-$5.23 & \bb{9.29} & $-$55.77 & $-$6.80 \\
\rowcolor{gray!10}
\methodname & \bb{6.19} & \bb{11.40} & \bb{$-$11.29} & \bb{$-$3.46} & \bb{9.29} & \bb{$-$54.92} & \bb{$-$5.54} \\
\bottomrule
\end{tabular}
\caption{WB-Scores on a 300-question subset of WildBench evaluated with GPT-5.5 as the judge.}
\label{tab:gpt55_judge}
\end{table*}

To address potential judge-dependent artifacts, we re-evaluated Vanilla and CoDIT for a subset of the teacher--student configurations reported in Table~\ref{tab:main_results} on a 300-question subset of WildBench, using gpt-5.5-2026-04-23 (\texttt{reasoning\_effort}: \texttt{high}) as a more advanced judge, selected for its significantly reduced hallucination rates~\citep{singh2026openaigpt5card}.

As shown in Table~\ref{tab:gpt55_judge}, in this rigorous evaluation setting, \methodname~matches or improves over the Vanilla baseline across all evaluated teacher--student configurations. These results indicate that the performance gains of \methodname~reflect genuine improvements in instruction-following capabilities.

\section{Dataset synthesis cost}
\label{app:synthesis_cost}

\begin{table*}[t]
\centering
\small
\begin{tabular}{l c cc cc c}
\toprule
\multirow{2}{*}{\textbf{Method}} & \multirow{2}{*}{\textbf{\#Responses}} & \multicolumn{2}{c}{\textbf{Teacher}} & \multicolumn{2}{c}{\textbf{Judge}} & \multirow{2}{*}{\textbf{Relative Cost}} \\
\cmidrule(lr){3-4} \cmidrule(lr){5-6}
 & & Time & Memory & Time & Memory & \\
\midrule
Vanilla & 1 & 940.8s & 91.70GB & -- & -- & 1.00$\times$ \\
Best-of-$N$ ($N=5$) & 5 & 940.8s & 91.70GB & 462.0s & 187.32GB & 10.02$\times$ \\
\rowcolor{gray!10}
\methodname & 1 & 1477.2s & 183.41GB & -- & -- & 3.14$\times$ \\
\bottomrule
\end{tabular}
\caption{Dataset synthesis cost measured on 5{,}000 instructions from LMSYS-Chat-1M with Qwen3-8B as the teacher model. Time denotes wall-clock time, and Memory denotes peak GPU memory consumption. Best-of-$N$ additionally requires judge annotation using gpt-oss-120b.}
\label{tab:synthesis_cost}
\end{table*}

The training cost of \methodname~is equivalent to that of Vanilla and Best-of-$N$, as all methods train the student model on the same number of instances. Since a synthesized dataset can be reused across a variety of student models once constructed, training cost is the primary practical concern. Nevertheless, we also report the dataset synthesis (decoding) cost in Table~\ref{tab:synthesis_cost}. The numbers are measured on 5{,}000 instructions from LMSYS-Chat-1M using Qwen3-8B as the teacher model. For Best-of-$N$, we assume $N=5$ independent runs, which additionally incur a judge annotation cost (gpt-oss-120b). Memory usage refers to peak GPU memory consumption.

\section{Evaluating the efficacy of instruction tuning}
\label{appendix:impact-of-post-training}

We evaluated our method using fully post-trained teacher models. However, post-training pipelines typically integrate instruction-tuning with subsequent alignment techniques, such as Direct Preference Optimization (DPO; \cite{rafailov2023direct}) or Reinforcement Learning (RL) variants, such as RLVR~\citep{lambert2025tulu, DBLP:journals/corr/abs-2501-12948,yue2025does}. To investigate whether the performance gains of \methodname~persist across these distinct training phases, we conduct an ablation study focusing on the teacher model's maturity.

\paragraph{Teacher Model}
We utilize the Olmo-3-7B-Instruct~\citep{olmo2025olmo3}, which provides intermediate checkpoints: the instruction-tuning-only model (Olmo-3-7B-Instruct-SFT) and the fully post-trained model after DPO and RLVR (Olmo-3-7B-Instruct). We generate synthetic datasets using each of these as a teacher. For all experiments, we fix $\alpha = 0.06$, using the same setting as Qwen3-8B.

\paragraph{Student Models}
To evaluate the effectiveness of the synthesized datasets, we employ Llama-3.1-8B~\citep{grattafiori2024llama3herdmodels} and Qwen3-8B-Base~\citep{yang2025qwen3technicalreport} as our student models. 

\paragraph{Datasets}
Following the data preparation process described in Section \ref{sec:experimental_setup}, we utilize the exact same set of 250,333 user instructions from the LMSYS-Chat-1M~\citep{zheng2024lmsyschatm} dataset to construct our synthetic training data.

\paragraph{Baselines}
For our comparative analysis, we adopt the \textit{Vanilla} baseline introduced in Section \ref{sec:experimental_setup}. 

\paragraph{Evaluation Datasets and Metrics}
Following the evaluation protocol described in Section \ref{sec:experimental_setup}, we assess the models' performance using AlpacaEval 2.0 and WildBench.

\paragraph{Results}

\begin{table*}[t]
\centering
\small
\setlength{\tabcolsep}{5pt}
\begin{tabular}{l ccc ccc}
\toprule
\multirow{3}{*}{\textbf{Datasets}} & \multicolumn{3}{c}{\textbf{Llama-3.1-8B}} & \multicolumn{3}{c}{\textbf{Qwen3-8B}} \\
\cmidrule(lr){2-4} \cmidrule(lr){5-7}
 & WildBench & \multicolumn{2}{c}{AlpacaEval 2.0} & WildBench & \multicolumn{2}{c}{AlpacaEval 2.0} \\
\cmidrule(lr){2-2} \cmidrule(lr){3-4} \cmidrule(lr){5-5} \cmidrule(lr){6-7}
 & WB-Score & LC & WR & WB-Score & LC & WR \\
\midrule
Vanilla (IT) &32.57 & 28.39&25.34 &39.84 & 35.79 & 31.88 \\
\rowcolor{gray!10}
\methodname (IT) &34.65 & 41.48	& 37.71 &44.35 & 46.43 & 41.02 \\
\midrule
Vanilla (IT + RL) & 47.96 & 41.01 & 49.20 &55.17 &47.00	&55.16\\
\rowcolor{gray!10}
\methodname (IT + RL) & \bb{50.04} & \bb{50.84} & \bb{58.79} & \bb{58.13} & \bb{54.48} & \bb{62.72} \\
\bottomrule
\end{tabular}
\caption{Performance comparison of student models (Llama-3.1-8B and Qwen3-8B-Base) trained on synthetic datasets generated by teacher models at different post-training stages (IT vs. IT + RL). LC stands for Length-Controlled win rate, and WR stands for Win Rate.}
\label{tab:post_training_stages_performance}
\end{table*}
The results of this evaluation are summarized in Table \ref{tab:post_training_stages_performance}. Based on these results, we make two key observations. First, the application of \methodname~consistently enhances the learning efficacy across both the instruction-tuned model and the fully post-trained model incorporating RL. This indicates that our approach is highly robust and can be consistently applied irrespective of the complex interactions and optimizations inherent in advanced post-training pipelines.
Second, the absolute performance of the student models is notably higher when distilled from the RL-aligned teacher compared to its instruction-tuning-only counterpart. We attribute this to the superior capabilities of the reinforcement-learned model, which intrinsically generates higher-quality and more refined responses, thereby providing a stronger learning signal for the student model.

\section{Additional results for chat vector distillation}
\label{appendix:additional_chat_vector_distillation}

\begin{figure}[t]
    \centering
    \includegraphics[width=0.5\linewidth]{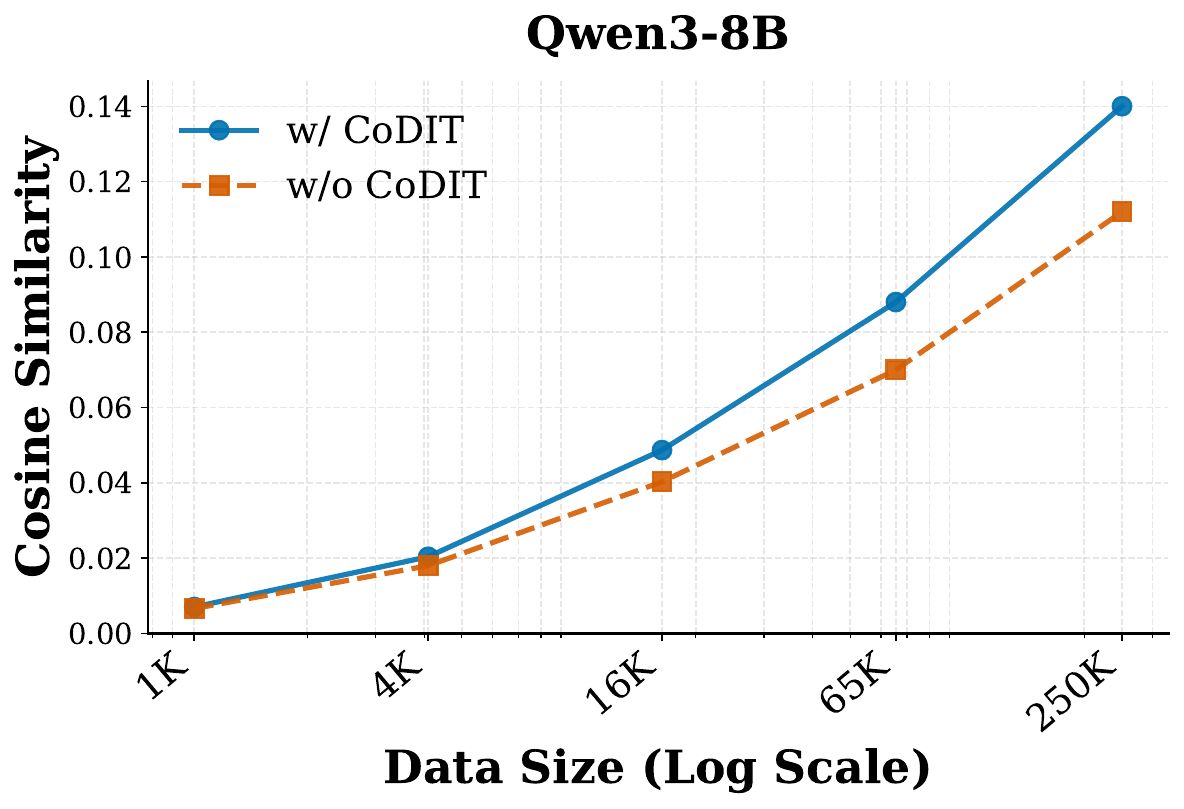}
    \caption{Cosine similarity between the parameter updates and the teacher's chat vector for Qwen3-8B. While the performance gain of \methodname~over the baseline is consistent with the trends observed in Llama models, the lower absolute similarity suggests that certain dimensions of the chat vector—specifically those associated with reasoning processes—are not fully captured, as thinking tokens were outside the scope of our current data construction.}
    \label{fig:cos_sims_qwen3}
\end{figure}

To verify whether the trends observed in Section~\ref{sec_chat_vector_distillation} hold for models with different architectures and training recipes, we conduct experiments using Qwen3-8B. 

As illustrated in Figure~\ref{fig:cos_sims_qwen3}, we confirm that Qwen3-8B~\citep{yang2025qwen3technicalreport} exhibits a similar trajectory to the Llama models: the cosine similarity between the induced update and the teacher's chat vector increases with the volume of training data, and the performance gap between \methodname~and the baseline progressively widens. 

However, we observe that the absolute cosine similarity for Qwen3-8B tends to be lower overall compared to the Llama series. We hypothesize that this is due to the nature of Qwen3-8B's post-training, which incorporates a thinking mode. Since its chat vector likely encompasses reasoning capabilities associated with these internal thought processes, and our current data construction does not include these reasoning (thinking) processes within its scope. Consequently, the similarity with the teacher's chat vector is lower than the values observed for the Llama models.

\end{document}